\documentclass[lettersize,journal]{IEEEtran}
\usepackage{amsmath,amsfonts}
\usepackage{algorithmic}
\usepackage{algorithm}
\usepackage{array}
\usepackage[caption=false,font=normalsize,labelfont=sf,textfont=sf]{subfig}
\usepackage{textcomp}
\usepackage{stfloats}
\usepackage{url}
\usepackage{verbatim}
\usepackage{graphicx}
\usepackage{cite}
\usepackage{soul}

\usepackage{amsmath}
\usepackage{booktabs}

\usepackage{verbatim}
\usepackage{textcomp}
\usepackage{stfloats}
\usepackage{algorithmic}
\usepackage{algorithm}

\usepackage[most]{tcolorbox}
\usepackage{xcolor}
\usepackage{lmodern}
\usepackage[T1]{fontenc}
\usepackage{multirow}
\usepackage{etoolbox}
\AtBeginEnvironment{tcolorbox}{\small}

\usepackage{booktabs}
\usepackage{listings}
\newcommand{\new}[1]{\textcolor{black}{#1}}

\newcommand{\hvvdata}{\texttt{HVVMemes}}
\newcommand{\hvvdatatwo}{\texttt{HVVMemes-2}}

\begin{document}

\title{Decoding Memes: Benchmarking Narrative Role Classification\\across Multilingual and Multimodal Models}


\author{Shivam Sharma,~\IEEEmembership{Member,~IEEE,}
        Tanmoy Chakraborty,~\IEEEmembership{{Senior Member,~IEEE}}%
\thanks{Manuscript received June 24, 2025.}%
\thanks{{Shivam Sharma and Tanmoy Chakraborty were with IIT Delhi. e-mail: shivam.sharma@ee.iitd.ac.in; tanmoy@ee.iitd.ac.in. Corresponding authors: Tanmoy Chakraborty. The work was partly supported by Anusandhan National Research Foundation
(CRG/2023/001351).}}
}



\maketitle

\begin{abstract}
This work investigates the challenging task of identifying narrative roles -- Hero, Villain, Victim, and Other -- in Internet memes, across three diverse test sets spanning English and code-mixed (English-Hindi) languages. Building on an annotated dataset originally skewed toward the `Other' class, we explore a more balanced and linguistically diverse extension, originally introduced as part of the CLEF 2024 shared task. Comprehensive lexical and structural analyses highlight the nuanced, culture-specific, and context-rich language used in real memes, in contrast to synthetically curated hateful content, which exhibits explicit and repetitive lexical markers. To benchmark the role detection task, we evaluate a wide spectrum of models, including fine-tuned multilingual transformers, sentiment and abuse-aware classifiers, instruction-tuned LLMs, and multimodal vision-language models. Performance is assessed under zero-shot settings using precision, recall, and F1 metrics. While larger models like DeBERTa-v3 and Qwen2.5-VL demonstrate notable gains, results reveal consistent challenges in reliably identifying the `Victim' class and generalising across cultural and code-mixed content. We also explore prompt design strategies to guide multimodal models and find that hybrid prompts incorporating structured instructions and role definitions offer marginal yet consistent improvements. Our findings underscore the importance of cultural grounding, prompt engineering, and multimodal reasoning in modelling subtle narrative framings in visual-textual content.\\
\textcolor{red}{\textbf{Warning:} This paper contains potentially harmful and offensive content.}
\end{abstract}


\section{Introduction}
Social media platforms have become pivotal arenas for rapid information dissemination. However, this openness has also catalysed the proliferation of harmful content -- including hate speech, propaganda, and misinformation, often embedded within \emph{memes} \cite{MacAvaney2019Hate, kiela2020hateful}. Memes, with their multimodal structure and cultural resonance, are particularly potent in shaping public opinion and propagating ideologies. While frequently humorous or satirical, memes can also encode malicious portrayals of individuals or groups, weaponising humour to veil defamation, bias, or misinformation \cite{mina2014batman, pramanick-etal-2021-detecting}.

\begin{figure}[t!]
    \centering
    \includegraphics[width=0.75\columnwidth]{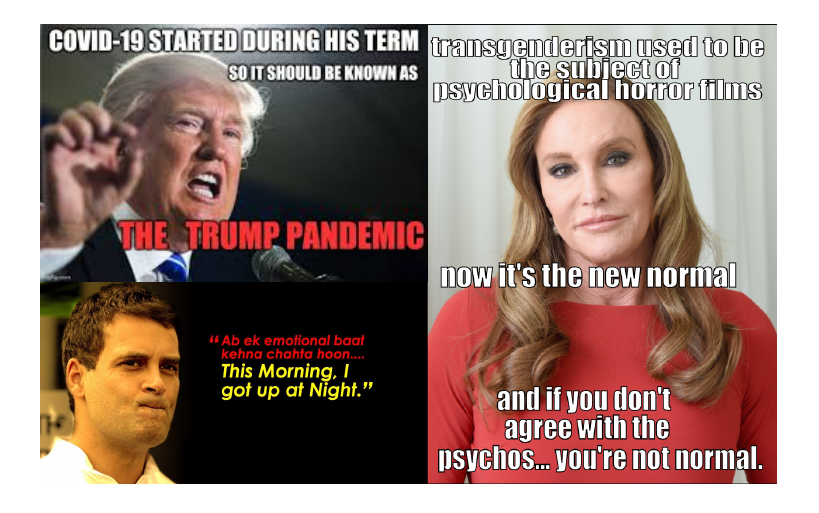}
    \caption{Examples of memes used, representing (a) \hvvdata~(En) \cite{sharma2022findings} - \textit{top-left}, (b) \hvvdatatwo~(En) \cite{clef2024sharedtask} -- \textit{bottom-left}, and (a) \hvvdatatwo~(EnHi) \cite{clef2024sharedtask} - \textit{right}.}
    \label{fig:intro_examples}
\end{figure}

\noindent\textbf{Motivation.}
Despite the growing attention toward multimodal meme analysis, previous research has predominantly addressed surface-level attributes such as hate speech, offensiveness, or sarcasm \cite{sharma-etal-2020-semeval, kumar2019sarc, zhou2021multimodal,memeEmotTAC}. However, a crucial yet relatively underexplored dimension is the \textit{semantic framing of entities}, specifically how memes depict individuals, communities, or institutions through narrative roles such as heroes, villains, victims, or others. These narrative portrayals transcend mere rhetorical devices and significantly influence audience perceptions, emotional reactions, and sociopolitical viewpoints. For instance, a meme attributing responsibility for a crisis to a public figure implicitly categorises them as a \emph{villain}, whereas memes expressing empathy or admiration portray entities respectively as \emph{victims} or \emph{heroes}.

\noindent\textbf{Challenges.}
Detecting entity roles in memes is a non-trivial task. Memes are replete with satirical, culturally grounded, and temporally situated content that often requires commonsense and visual reasoning to decode. The connotation of an entity is rarely explicit and frequently context-dependent, demanding reasoning beyond simple keyword matching or sentiment classification. Moreover, code-mixed memes add another layer of complexity due to transliteration, idiomatic phrasing, and socio-cultural references \cite{DISARM:2022,sharma-etal-2023-characterizing}.

Figure~\ref{fig:intro_examples} illustrates such narratives. One meme vilifies Donald Trump by attributing the COVID-19 outbreak to his tenure. Another ridiculous Rahul Gandhi’s speech to delegitimise his political credibility. A third mocks the visibility of trans people in modern media by juxtaposing them with historical horror tropes. These portrayals underscore the power of memes to encode complex, often contentious, societal narratives.

Prior studies \cite{sharma2022findings, sharma-etal-2023-characterizing} have ventured into examining narrative portrayals within realistic meme contexts, yet their investigations remain restricted to a narrow set of domains, specifically US Politics and COVID-19, and exclusively in English. This limitation overlooks significant insights available from other linguistic and culturally nuanced environments, particularly low-resource yet widely used code-mixed settings involving languages such as English and Hindi (Hinglish). While recent works  \cite{semeval2025task10} delve deeper into narrative roles across multiple languages, their analyses remain limited to textual modalities.



Building on prior work introduced in CONSTRAINT-2022 \cite{sharma2022findings}, we revisit the challenge of detecting semantic roles in memes -- \textit{hero}, \textit{villain}, \textit{victim}, and \textit{other} -- from a multimodal and multilingual perspective. We extensively benchmark the first iteration of the \hvvdata~dataset and introduce additional evaluations in two language settings: English and code-mixed (English-Hindi), for the task introduced as part of the CLEF 2024 CheckThat! Lab \cite{clef2024sharedtask}. These additions address resource-constrained and culturally nuanced settings, revealing challenges in role recognition across languages and contexts.

Unlike existing models that often rely on ensemble-based or template-matching strategies \cite{Logically:2022, DD-TIG:2022}, our study investigates the comparative performance of over 25 models -- including sentiment-tuned, multilingual, and vision-language architectures on nuanced narrative understanding. We also explore a suite of prompt-engineering strategies to assess their effectiveness in improving model performance.


\noindent\textbf{Our Contributions.}
To address these challenges, we present a comprehensive study on narrative role detection in memes, with the following key contributions:

\begin{itemize}
    \item \textbf{Dataset Analysis:} Lexical and structural analyses across real vs. synthetic and English vs. Hinglish test sets reveal cultural and stylistic differences in meme construction.

    \item \textbf{Prompt Engineering:} Among four variants, a hybrid prompt (P4) combining structural cues and role definitions best generalises across roles, especially \emph{villain} and \emph{hero}.

    \item \textbf{Model Benchmarking:} We evaluate 25+ models (sentiment-tuned, instruction-following LLMs, multimodal), finding that smaller or narrowly trained models struggle with subtle role distinctions.

    \item \textbf{Cross-Scenario Evaluation:} Zero-shot results across test sets show inconsistent generalisation, with \emph{victim} and \emph{other} roles being particularly challenging.

    \item \textbf{Qualitative Insights:} Analysis of representative predictions shows vision-language models excel in nuanced meme interpretation via multimodal alignment.

    \item \textbf{Embedding \& Error Analysis:} Embedding visualisations and confusion matrices expose semantic ambiguities and cultural heterogeneity in narrative role modeling.
\end{itemize}

\section{Related Work}

\paragraph{\bf Online Harmfulness}
With the rapid proliferation of harmful content on social media platforms, extensive research has emerged exploring various aspects of online harmfulness. These studies encompass online trolling, cyber-bullying, cyber-stalking, and hate speech \cite{Survey:2022:Harmful:Memes,hee-etal-2024-recent}. Additionally, investigations have highlighted connections between online racial and ethnic discrimination and corresponding offline experiences \cite{Relia_Li_Cook_Chunara_2019}. Psychological and sociological perspectives on trolling behaviours were examined in \cite{chengtroll2017}. Other notable studies have analysed homophily regarding self-harm linked to eating disorders using logistic regression and snowball sampling, as well as suicide ideation employing linguistic, structural, affective, and socio-psychological features \cite{Survey:2022:Harmful:Memes,IJCAI2020:propaganda:survey}. However, much of this prior research has predominantly focused on text-based analysis, leaving other modalities underexplored.

\paragraph{\bf Characterising Online Targets}
Another pertinent research stream investigates aspects such as relevance, stance, hate speech, sarcasm, and dialogue acts within hateful exchanges, specifically on Twitter, employing conventional and multi-task settings \cite{Gautam_Mathur_Gosangi_Mahata_Sawhney_Shah_2020,ousidhoum-etal-2019-multilingual}. A few works also prescribe neural network-based approaches with word embeddings and aspect-based sentiment analysis, while tackling challenges like data sparsity, accuracy, and sarcastic content identification \cite{zain2018neural}. Additionally, authors in \cite{shvets-etal-2021-targets} highlighted the utility of a generic concept extraction module for hate speech target identification. Further studies have addressed social biases and explainability regarding targeted protected groups \cite{mathew2020hatexplain,sap-etal-2020-social}. Hierarchical bidirectional gated recurrent units were utilised \cite{ma-etal-2018-joint} for detecting targets and associated sentiments, while sequence tagging approaches in low-resource contexts were adopted in \cite{mitchell-etal-2013-open}. 
Most studies typically focus on singular designated targets or sentiment detection, overlooking broader affective nuances, limiting generalisation across domains as noted in \cite{DISARM:2022,shvets-etal-2021-targets}.

\paragraph{\bf Studies on Memes}
The influx of memes from online fringe communities such as Gab, Reddit, and 4chan to mainstream platforms like Twitter and Instagram has led to widespread dissemination of intentionally harmful content \cite{Zannettou2018}. Investigations into harmful multimodal meme detection have employed conventional visual features combined with multimodal associativity \cite{CURwebist20,memenonmeme}. Multiple datasets have been curated to capture offensive \cite{suryawanshi-etal-2020-multimodal}, hateful \cite{kiela2020hateful,gomez2019exploring}, and harmful content \cite{pramanick-etal-2021-momenta-multimodal}. Specific research has focused on memetic harmfulness detection and targeted categories \cite{pramanick-etal-2021-momenta-multimodal}. Additionally, commonsense knowledge \cite{9582340}, web entities, racial cues \cite{pramanick-etal-2021-momenta-multimodal,karkkainen2019fairface,sharma-etal-2023-memex,agarwal-etal-2024-mememqa}, and other external information have been explored for identifying offensive, harmful, and hateful memes. 
Most approaches typically either focus on general harmfulness detection tasks  \cite{Survey:2022:Harmful:Memes}, rely on ensemble methodologies lacking optimal cost efficiency, or do not consider related tasks within the context of cross-lingual examination. To date, no standalone solution robustly addresses the nuanced task of identifying specific roles of entities portrayed within memes in a balanced way. This gap underscores the necessity for an exhaustive benchmarking, encompassing the mix of conventional and recent advances in suitable modelling strategies, and thereby decoding the challenges underlying the complex task.
\section{Dataset}

\begin{table}[t!]
\centering
\caption{Role-wise counts of the entity-references as different roles in \hvvdata~\cite{sharma2022findings} and \hvvdatatwo~\cite{clef2024sharedtask} datasets.}
\label{tab:stat_hvvp2}
\begin{tabular}{cccc}
\toprule
\textbf{Roles} &  \textbf{EN} &  \textbf{EN} & \textbf{EN-HI}  \\
\midrule
Hero           & 49 & 144            & 252          \\
Villain        & 273 & 404            & 348          \\
Victim         & 95 & 122            & 207          \\
Other          & 667 & 141            & 148         \\
\textbf{Total} & \textbf{1084} & \textbf{811}   & \textbf{955}\\
\bottomrule
\end{tabular}%
\end{table}

\new{The dataset introduced in \cite{sharma2022findings} incorporated narrative role labels spanning four categories: \textit{hero, villain, victim, and other}. The distribution of entity references across these roles was notably skewed, with 49 heroes, 273 villains, 95 victims, and a substantial 667 categorised as other, highlighting a dominance of the `other' category. The data primarily comprised realistic memes centred on US politics and COVID-19.}

\new{Subsequent evaluations utilised two distinct language settings: one in English, and the other featuring code-mixed English memes. These were released as part of the CLEF 2024 shared task \cite{clef2024sharedtask}. In contrast to the earlier dataset, both the English and code-mixed test sets featured a fairly balanced distribution of entity roles. As detailed in Table \ref{tab:stat_hvvp2}, the English test set comprised 811 samples, while the code-mixed counterpart included 955 samples, offering a more representative spread across narrative roles. The statistical summary of these datasets can be found in Table \ref{tab:stat_hvvp2}.}


\begin{figure*}[t!]
    \centering
    \includegraphics[width=0.85\linewidth]{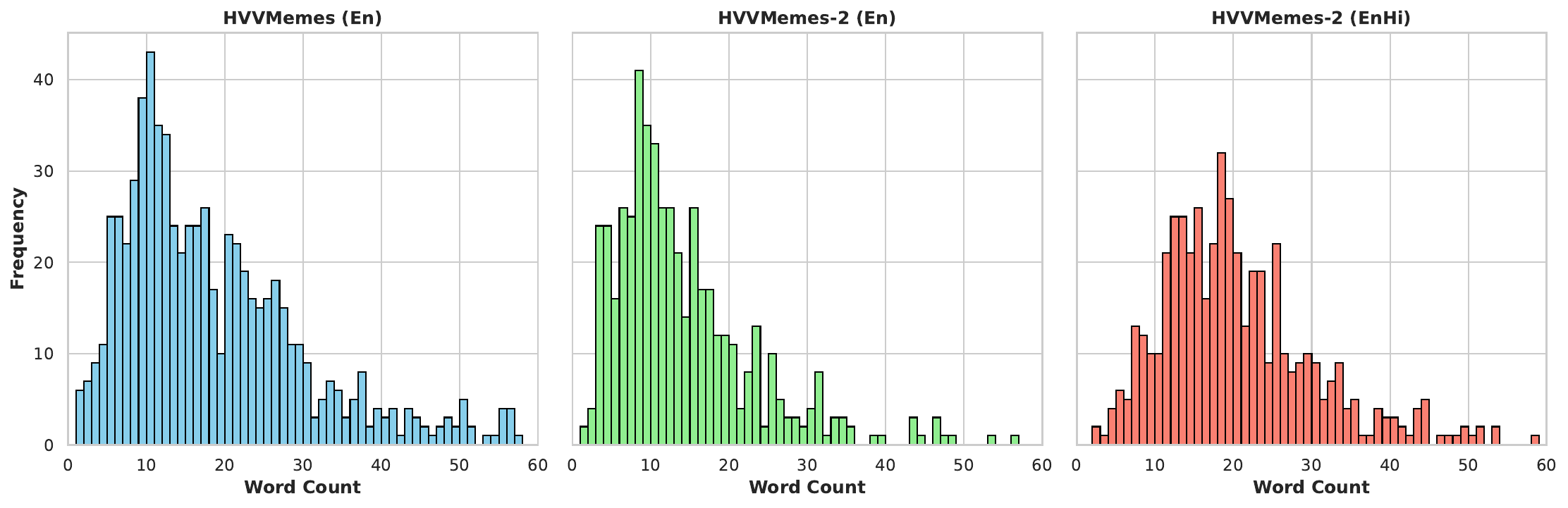}
    \caption{Embedded text length distribution for \hvvdata\ (En) -- \textit{left}, \hvvdatatwo~(En) -- \textit{middle} and \hvvdatatwo~(EnHi) -- \textit{right}.}
    \label{fig:ocr_dist}
\end{figure*}

\paragraph{Word Length Distribution Across Test Sets}
\new{This analysis focuses on the word length distribution of meme text across the test sets -- \hvvdata~(En), \hvvdatatwo~(En) and \hvvdatatwo~(EnHi), shown in Figure \ref{fig:ocr_dist}. The first and second sets, \hvvdata~(En) and \hvvdatatwo~(En), comprising English-language memes, exhibit similar patterns: most memes contain 10 to 15 words, with the overall distribution being left-skewed and gradually tapering off towards 60 words. This suggests a preference for concise, pop-culture-driven memes. In contrast, the third test set, consisting of Hinglish code-mixed memes from the Indian context, displays a broader and more uniform distribution, mainly between 10 to 25 words. This indicates greater variation in meme text length, styles, and templates, highlighting cultural and lexical diversity in meme creation and communication} (see Appendix A in supplementary for more details on lexical analysis of the test sets).
\section{Experimental Details}


\paragraph{Baselines}To establish comparative benchmarks, we evaluate a diverse set of baseline models spanning multilingual language models, domain-specific variants, Hinglish-focused sentiment models, and recent instruction-tuned and multimodal architectures. These baselines reflect both prior state-of-the-art systems and emerging capabilities relevant to our task (see Appendix A in supplementary for more details on comparative baselines used).

\begin{table*}[t!]
\centering
\caption{Results from prompt configurations explored across different test sets via the LLaVA-NeXT model.}
\label{tab:prompt}
\resizebox{\textwidth}{!}{%
\begin{tabular}{cccccccccccccccc}
\hline
\multirow{2}{*}{\textbf{Prompts}} & \multicolumn{3}{c}{\textbf{Hero (52)}} & \multicolumn{3}{c}{\textbf{Other (1917)}} & \multicolumn{3}{c}{\textbf{Victim (114)}} & \multicolumn{3}{c}{\textbf{Villain (350)}} & \multicolumn{3}{c}{\textbf{Macro (2433)}} \\ \cline{2-16} 
 & Prec & Rec & F1 & Prec & Rec & F1 & Prec & Rec & F1 & Prec & Rec & F1 & Prec & Rec & F1 \\ \hline
\multicolumn{16}{c}{\textbf{\hvvdata~(En)}} \\ \hline
P4 & 0.096 & 0.269 & 0.141 & 0.917 & 0.294 & 0.445 & 0.250 & 0.035 & 0.062 & 0.190 & 0.900 & 0.314 & 0.363 & 0.375 & 0.240 \\
P3 & 0.133 & 0.115 & 0.124 & 0.884 & 0.695 & 0.778 & 1.000 & 0.018 & 0.034 & 0.276 & 0.691 & 0.394 & 0.573 & 0.380 & 0.333 \\ \hline
\multicolumn{16}{c}{\textbf{\hvvdatatwo~(En)}} \\ \hline
P3 & 0.444 & 0.028 & 0.052 & 0.246 & 0.723 & 0.367 & 1.000 & 0.016 & 0.032 & 0.527 & 0.502 & 0.515 & 0.554 & 0.318 & 0.242 \\
P4 & 0.387 & 0.083 & 0.137 & 0.236 & 0.390 & 0.294 & 0.550 & 0.090 & 0.155 & 0.524 & 0.683 & 0.593 & 0.424 & 0.312 & 0.295 \\ \hline
\multicolumn{16}{c}{\textbf{\hvvdatatwo~(En-Hi)}} \\ \hline
P3 & 0.561 & 0.091 & 0.157 & 0.166 & 0.872 & 0.279 & 1.000 & 0.005 & 0.010 & 0.426 & 0.167 & 0.240 & 0.538 & 0.284 & 0.171 \\
P2 & 0.507 & 0.139 & 0.218 & 0.170 & 0.905 & 0.287 & 0.500 & 0.005 & 0.010 & 0.567 & 0.158 & 0.247 & 0.436 & 0.302 & 0.190 \\
P1 & 0.314 & 0.456 & 0.372 & 0.000 & 0.000 & 0.000 & 0.182 & 0.010 & 0.018 & 0.375 & 0.621 & 0.468 & 0.218 & 0.272 & 0.215 \\
P4 & 0.372 & 0.266 & 0.310 & 0.154 & 0.392 & 0.221 & 0.300 & 0.014 & 0.028 & 0.373 & 0.417 & 0.393 & 0.300 & 0.272 & 0.238 \\ \hline
\end{tabular}%
}
\end{table*}

\paragraph{Evaluation Setup}

As part of benchmarking \hvvdata~and \hvvdatatwo, we evaluate the performance of three main model categories -- smaller text-only language models, text-only LLMs, and multimodal LLMs -- on the task of identifying heroes, villains, and victims in memes. This evaluation spans three scenarios: (a) English memes related to US Politics and COVID-19 from the \hvvdata~test set, and English as well as code-mixed (English-Hindi) memes covering a broad range of topics from the \hvvdatatwo~test sets. Being a multi-class classification scenario per sample, the results are evaluated across different scenarios using conventional metrics: precision, recall, and F1-score across four classes -- Hero, Victim, Villain, and Other -- along with their macro-average.

\section{Prompting Configuration}

To investigate the impact of prompt design on response quality, we explore four distinct prompting strategies using LLaVA-NeXT\footnote{See Appendix A in supplementary for more details on the prompting configurations.}, first on \hvvdatatwo~(En-Hi), followed by \hvvdatatwo~(En) and \hvvdata. We choose to start from the most challenging setting first in order to explore the configuration that generalises well across scenarios. The first prompting approach employs a basic, unoptimized prompt serving as a baseline (P1). The second strategy introduces an explicit emphasis on the \textit{other} category to reduce ambiguity in classification (P2). The third configuration adopts an optimised prompt, incorporating clear definitions for each category alongside formatting instructions to enhance structure and clarity (P3). Finally, the hybrid prompt combines narrative-role label definitions with a custom system instruction (combining first and third prompts), aiming to balance precision with flexibility (P4). These variations allow us to systematically evaluate how different levels of prompt specificity influence the model's interpretability and output consistency. The results are reported in Table \ref{tab:prompt}.
 
Interestingly, with P1 (a basic prompt), the F1 score is higher at 0.21 compared to 0.19 for the scenario when explicitly prompted to emphasise the usage instruction for the  \textit{other} category -- P2. The second scenario is examined when no true positives for the \textit{other} category for the baseline prompt are encountered (cf. Table \ref{tab:prompt}). Further optimising the prompt and using role definitions does not yield an overall better performance than observed, especially for the category \textit{victim}; however, it preserves the prediction performance across other categories. This suggests the utility of including the definitions (while not explicitly asking to emphasise any), providing more guidance to the task. 
Additionally, prompts P2 and P3 have slightly higher precision, recall and overall score as compared to P4 for the \textit{other} category. However, they fall behind due to the difference in role detection for the remaining categories.  

With the latest modifications (P4), the prompt now serves as a hybrid of the basic scenario (P1) and the version that included definitions of the role labels (P3). This leads to a slight overall performance increase. The performance of the \textit{hero} category is observed to improve compared to the scenario with emphasis on role definitions and structured output instruction, though it does not perform as well for the other category. The F1 score increases from 21 to 31, which is slightly lower than the basic prompt but maintains the \textit{other} category with a 0.22 F1 score instead of a 0 F1 score—a significant improvement. The \textit{victim} category shows a marginal increase in performance, though it remains very poor and requires further attention. The performance for the \textit{villain} category also improves, rising from 0.24 to 0.39 F1 score, which is a notable positive development. Overall, the F1 score increases from 0.21, which is previously the highest achieved with the basic prompt (P1), to 0.23 with this new hybrid prompt (P4). However, the challenge of identifying \textit{victims} persists. In summary, there is a noticeable enhancement from the definition-based prompt to the hybrid one across the categories \textit{hero}, \textit{villain}, and overall, while the \textit{other} category remains stable with a slight reduction, and the \textit{victim} category is still performing poorly.
\begin{table*}[t!]
\centering
\caption{Benchmarking results on the HVV Phase 1 test set.}
\label{tab:hvvp1results}
\resizebox{\textwidth}{!}{%
\begin{tabular}{lccccccccccccccc}
\toprule
 & \multicolumn{3}{c}{\textbf{Hero (52)}} & \multicolumn{3}{c}{\textbf{Other (1917)}} & \multicolumn{3}{c}{\textbf{Victim (114)}} & \multicolumn{3}{c}{\textbf{Villain (350)}} & \multicolumn{3}{c}{\textbf{Macro (2433)}} \\
 \cmidrule{2-16}
\multicolumn{1}{c}{\textbf{}} & Prec & Rec & F1 & Prec & Rec & F1 & Prec & Rec & F1 & Prec & Rec & F1 & Prec & Rec & F1 \\
\midrule
{\color[HTML]{9B9B9B} Majority} & {\color[HTML]{9B9B9B} 0.000} & {\color[HTML]{9B9B9B} 0.000} & {\color[HTML]{9B9B9B} 0.000} & {\color[HTML]{9B9B9B} 0.790} & {\color[HTML]{9B9B9B} 1.000} & {\color[HTML]{9B9B9B} 0.880} & {\color[HTML]{9B9B9B} 0.000} & {\color[HTML]{9B9B9B} 0.000} & {\color[HTML]{9B9B9B} 0.000} & {\color[HTML]{9B9B9B} 0.000} & {\color[HTML]{9B9B9B} 0.000} & {\color[HTML]{9B9B9B} 0.000} & {\color[HTML]{9B9B9B} 0.200} & {\color[HTML]{9B9B9B} 0.250} & {\color[HTML]{9B9B9B} 0.220} \\
{\color[HTML]{9B9B9B} Random} & {\color[HTML]{9B9B9B} 0.030} & {\color[HTML]{9B9B9B} 0.370} & {\color[HTML]{9B9B9B} 0.060} & {\color[HTML]{9B9B9B} 0.770} & {\color[HTML]{9B9B9B} 0.250} & {\color[HTML]{9B9B9B} 0.380} & {\color[HTML]{9B9B9B} 0.040} & {\color[HTML]{9B9B9B} 0.200} & {\color[HTML]{9B9B9B} 0.060} & {\color[HTML]{9B9B9B} 0.170} & {\color[HTML]{9B9B9B} 0.320} & {\color[HTML]{9B9B9B} 0.220} & {\color[HTML]{9B9B9B} 0.250} & {\color[HTML]{9B9B9B} 0.280} & {\color[HTML]{9B9B9B} 0.180} \\
\midrule
\multicolumn{16}{c}{\textbf{Text-Only Models}} \\
\midrule
muril-base-cased & 0.333 & 0.019 & 0.036 & 0.815 & 0.967 & 0.884 & 0.270 & 0.088 & 0.132 & 0.483 & 0.163 & 0.244 & 0.475 & 0.309 & 0.324 \\

bert-base (ml-cm-c-sent) & 0.212 & 0.135 & 0.165 & 0.827 & 0.877 & 0.852 & 0.148 & 0.412 & 0.218 & 0.510 & 0.071 & 0.125 & 0.424 & 0.374 & 0.340 \\
 bert-base (ml-c) & 0.167 & 0.019 & 0.034 & 0.824 & 0.877 & 0.850 & 0.330 & 0.281 & 0.303 & 0.233 & 0.191 & 0.210 & 0.388 & 0.342 & 0.349 \\

bert-base (ml-uc-sent) & 0.000 & 0.000 & 0.000 & 0.852 & 0.852 & 0.852 & 0.320 & 0.140 & 0.195 & 0.323 & 0.426 & 0.367 & 0.374 & 0.354 & 0.354 \\

hinglish11k-sentiment-analysis & 0.063 & 0.269 & 0.101 & 0.848 & 0.841 & 0.844 & 0.234 & 0.377 & 0.289 & 0.460 & 0.163 & 0.241 & 0.401 & 0.413 & 0.369 \\
gk-hinglish-sentiment & 0.136 & 0.115 & 0.125 & 0.836 & 0.870 & 0.853 & 0.207 & 0.211 & 0.209 & 0.361 & 0.286 & 0.319 & 0.385 & 0.370 & 0.376 \\

OGBV-gender-indicbert (cm) & 0.028 & 0.019 & 0.023 & 0.846 & 0.884 & 0.864 & 0.278 & 0.193 & 0.228 & 0.416 & 0.374 & 0.394 & 0.392 & 0.368 & 0.377 \\

hindi-abusive-MuRIL (cm) & 0.107 & 0.115 & 0.111 & 0.899 & 0.773 & 0.831 & 0.330 & 0.325 & 0.327 & 0.354 & 0.626 & 0.452 & 0.423 & 0.460 & 0.431 \\
muril-base (c-ft-cm) & 0.190 & 0.212 & 0.200 & 0.902 & 0.745 & 0.816 & 0.339 & 0.342 & 0.341 & 0.332 & 0.643 & 0.438 & 0.441 & 0.485 & 0.449 \\
muril-large (c) & 0.184 & 0.135 & 0.156 & 0.886 & 0.839 & 0.862 & 0.381 & 0.421 & 0.400 & 0.434 & 0.560 & 0.489 & 0.471 & 0.489 & 0.477 \\
deberta-v3-large & 0.281 & 0.308 & 0.294 & 0.891 & 0.863 & 0.877 & 0.443 & 0.474 & 0.458 & 0.514 & 0.583 & 0.546 & 0.532 & 0.557 & 0.543 \\
\midrule
Llama-3.1-8B-Instruct & 0.080 & 0.080 & 0.080 & 0.910 & 0.260 & 0.400 & 0.090 & 0.010 & 0.020 & 0.180 & 0.930 & 0.300 & 0.320 & 0.320 & 0.200 \\
phi-2 & 0.020 & 0.230 & 0.030 & 0.780 & 0.690 & 0.730 & 0.400 & 0.020 & 0.030 & 0.330 & 0.050 & 0.080 & 0.380 & 0.250 & 0.220 \\
Mistral-7B-Instruct-v0.1 & 0.020 & 0.040 & 0.030 & 0.800 & 0.850 & 0.820 & 0.070 & 0.150 & 0.100 & 0.440 & 0.090 & 0.150 & 0.330 & 0.280 & 0.280 \\
Qwen2.5-7B-Instruct & 0.090 & 0.120 & 0.100 & 0.860 & 0.710 & 0.780 & 0.210 & 0.300 & 0.250 & 0.290 & 0.510 & 0.370 & 0.360 & 0.410 & 0.370 \\
\midrule
\multicolumn{16}{c}{\textbf{Multimodal Models}} \\
\midrule
Qwen2-VL-7B-Instruct & 0.090 & 0.500 & 0.160 & 0.890 & 0.020 & 0.030 & 0.100 & 0.450 & 0.170 & 0.190 & 0.890 & 0.310 & 0.320 & 0.460 & 0.170 \\
CLIP (raw text) & 0.000 & 0.000 & 0.000 & 0.787 & 0.993 & 0.878 & 0.000 & 0.000 & 0.000 & 0.125 & 0.006 & 0.011 & 0.228 & 0.250 & 0.222 \\
LLaVA-NeXT (P4) & 0.096 & 0.269 & 0.141 & 0.917 & 0.294 & 0.445 & 0.250 & 0.035 & 0.062 & 0.190 & 0.900 & 0.314 & 0.363 & 0.375 & 0.240 \\
CLIP & 0.000 & 0.000 & 0.000 & 0.797 & 0.948 & 0.866 & 0.135 & 0.184 & 0.156 & 0.000 & 0.000 & 0.000 & 0.233 & 0.283 & 0.256 \\
Qwen2.5-VL-7B-Instruct & 0.110 & 0.270 & 0.160 & 0.920 & 0.280 & 0.430 & 0.100 & 0.480 & 0.170 & 0.250 & 0.830 & 0.380 & 0.340 & 0.470 & 0.280 \\
LLaVA-NeXT (P3) & 0.133 & 0.115 & 0.124 & 0.884 & 0.695 & 0.778 & 1.000 & 0.018 & 0.034 & 0.276 & 0.691 & 0.394 & 0.573 & 0.380 & 0.333\\
\bottomrule
\end{tabular}%
}
\end{table*}

\section{Benchmarking Insights}

We present a comparative analysis of the performance of various models across different narrative roles and test scenarios, highlighting key trends, model strengths, and persistent challenges, particularly in detecting nuanced or underrepresented roles.

\subsection{Performance on \hvvdata~(En) test set}

\paragraph{Traditional Language Models (Fine-tuned)}
Table \ref{tab:hvvp1results} suggests that the performance of traditional models, including BERT-based multilingual variants and sentiment-tuned models, is generally suboptimal. However, code-mixed models trained on sentiment data exhibit relatively better overall performance (F1 = 0.35). This effect is particularly observed for the \textit{villain} category. 
The highest macro F1 in this group is observed for deberta-v3-large (F1 = 0.54). This is closely followed by different \text{muril} variants, outperforming other multilingual baselines.
Notably, all models in this group achieve their highest performance in the \textit{other} category (F1 = $0.81-0.88$), reflecting dataset imbalance, as also corroborated by the majority baseline of $0.88$. These observations offer valuable insights into (a) the effectiveness of adapting a multilingual model for monolingual tasks, and (b) the common trend of performance scaling with model size.

\paragraph{Text-only Large Language Models (Zero-shot)}
To evaluate inherent reasoning abilities, we test LLMs in a zero-shot setting. LLaMA 3.1 8B Instruct and Phi-2 achieve low macro F1-scores (0.20-0.22). Interestingly, LLaMA 3.1 shows the second-best F1 for the \textit{villain} category, possibly due to the strong semantic correlations for vilifying narratives induced by its general-purpose understanding capabilities, whereas Phi-2, having been trained over synthetic data, exhibits limitations in analysing context with subtle nuances. Mistral-7B-Instruct-v0.1, although offering marginal improvements, performed well on \textit{other}.  However, Qwen2.5-7B-Instruct is observed to yield best performance in this category with a macro F1 of 0.37 -- underscoring strong performances for \textit{other} (F1 = 0.78) and \textit{villain} (F1 = 0.37) likely due to better multi-lingual and complex reasoning capabilities. Overall, the performances remain sub-optimal for the \textit{hero} role and moderate for \textit{victim}, with Qwen2.5-7B-Instruct achieving the comparatively highest F1 scores of 0.10 and 0.25, respectively. In contrast, deberta-v3-large attains a significantly higher overall score of 0.54, highlighting a considerable gap in the reasoning capabilities of current LLMs.
\begin{table*}[t!]
\centering
\caption{Benchmarking results on the HVV Phase 2 test set - En.}
\label{tab:hvvp2Enresults}
\resizebox{\textwidth}{!}{%
\begin{tabular}{@{}lccccccccccccccc@{}}
\toprule
 & \multicolumn{3}{c}{\textbf{Hero (144)}} & \multicolumn{3}{c}{\textbf{Other (141)}} & \multicolumn{3}{c}{\textbf{Victim (122)}} & \multicolumn{3}{c}{\textbf{Villain (404)}} & \multicolumn{3}{c}{\textbf{Macro (811)}} \\ \cmidrule(l){2-16} 
\textbf{} & \textbf{Prec} & \textbf{Rec} & \textbf{F1} & \textbf{Prec} & \textbf{Rec} & \textbf{F1} & \textbf{Prec} & \textbf{Rec} & \textbf{F1} & \textbf{Prec} & \textbf{Rec} & \textbf{F1} & \textbf{Prec} & \textbf{Rec} & \textbf{F1} \\ \midrule
{\color[HTML]{9B9B9B} Majority} & {\color[HTML]{9B9B9B} 0.000} & {\color[HTML]{9B9B9B} 0.000} & {\color[HTML]{9B9B9B} 0.000} & {\color[HTML]{9B9B9B} 0.000} & {\color[HTML]{9B9B9B} 0.000} & {\color[HTML]{9B9B9B} 0.000} & {\color[HTML]{9B9B9B} 0.000} & {\color[HTML]{9B9B9B} 0.000} & {\color[HTML]{9B9B9B} 0.000} & {\color[HTML]{9B9B9B} 0.500} & {\color[HTML]{9B9B9B} 1.000} & {\color[HTML]{9B9B9B} 0.670} & {\color[HTML]{9B9B9B} 0.120} & {\color[HTML]{9B9B9B} 0.250} & {\color[HTML]{9B9B9B} 0.170} \\
{\color[HTML]{9B9B9B} Random} & {\color[HTML]{9B9B9B} 0.150} & {\color[HTML]{9B9B9B} 0.200} & {\color[HTML]{9B9B9B} 0.170} & {\color[HTML]{9B9B9B} 0.160} & {\color[HTML]{9B9B9B} 0.240} & {\color[HTML]{9B9B9B} 0.190} & {\color[HTML]{9B9B9B} 0.120} & {\color[HTML]{9B9B9B} 0.190} & {\color[HTML]{9B9B9B} 0.150} & {\color[HTML]{9B9B9B} 0.530} & {\color[HTML]{9B9B9B} 0.280} & {\color[HTML]{9B9B9B} 0.370} & {\color[HTML]{9B9B9B} 0.240} & {\color[HTML]{9B9B9B} 0.230} & {\color[HTML]{9B9B9B} 0.220} \\ \midrule
\multicolumn{16}{c}{\textbf{Text-Only Models}} \\ \midrule
muril-base-cased & 0.500 & 0.007 & 0.014 & 0.188 & 0.979 & 0.315 & 0.220 & 0.090 & 0.128 & 0.708 & 0.042 & 0.079 & 0.404 & 0.279 & 0.134 \\

bert-base (ml-cm-c-sent) & 0.000 & 0.000 & 0.000 & 0.239 & 0.794 & 0.368 & 0.192 & 0.525 & 0.281 & 0.500 & 0.007 & 0.015 & 0.233 & 0.332 & 0.166 \\
gk-hinglish-sentiment & 0.125 & 0.007 & 0.013 & 0.231 & 0.908 & 0.368 & 0.180 & 0.254 & 0.211 & 0.618 & 0.116 & 0.196 & 0.289 & 0.321 & 0.197 \\
OGBV-gender-indicbert (cm) & 0.250 & 0.014 & 0.026 & 0.215 & 0.915 & 0.348 & 0.219 & 0.205 & 0.212 & 0.562 & 0.124 & 0.203 & 0.312 & 0.314 & 0.197 \\

hinglish11k-sentiment-analysis & 0.229 & 0.111 & 0.150 & 0.256 & 0.809 & 0.389 & 0.167 & 0.393 & 0.234 & 0.750 & 0.015 & 0.029 & 0.350 & 0.332 & 0.200 \\
 bert-base (ml-c) & 0.000 & 0.000 & 0.000 & 0.241 & 0.865 & 0.377 & 0.236 & 0.205 & 0.219 & 0.629 & 0.307 & 0.413 & 0.276 & 0.344 & 0.252 \\

bert-base (ml-uc-sent) & 0.250 & 0.007 & 0.014 & 0.252 & 0.823 & 0.386 & 0.215 & 0.115 & 0.150 & 0.606 & 0.423 & 0.499 & 0.331 & 0.342 & 0.262 \\

hindi-abusive-MuRIL (cm) & 0.184 & 0.063 & 0.093 & 0.260 & 0.723 & 0.382 & 0.199 & 0.320 & 0.245 & 0.595 & 0.255 & 0.357 & 0.309 & 0.340 & 0.269 \\
muril-large (c) & 0.333 & 0.014 & 0.027 & 0.269 & 0.816 & 0.404 & 0.235 & 0.344 & 0.279 & 0.626 & 0.307 & 0.412 & 0.366 & 0.370 & 0.280 \\
deberta-v3-large & 0.250 & 0.035 & 0.061 & 0.256 & 0.908 & 0.399 & 0.285 & 0.385 & 0.328 & 0.706 & 0.220 & 0.336 & 0.374 & 0.387 & 0.281 \\
muril-base (c-ft-cm) & 0.294 & 0.069 & 0.112 & 0.305 & 0.759 & 0.435 & 0.212 & 0.344 & 0.263 & 0.583 & 0.329 & 0.421 & 0.349 & 0.375 & 0.308 \\ \midrule
phi-2 & 0.160 & 0.220 & 0.180 & 0.160 & 0.650 & 0.250 & 0.140 & 0.010 & 0.020 & 0.500 & 0.020 & 0.030 & 0.240 & 0.220 & 0.120 \\
Mistral-7B-Instruct-v0.1 & 0.170 & 0.100 & 0.130 & 0.190 & 0.790 & 0.300 & 0.160 & 0.130 & 0.140 & 0.410 & 0.020 & 0.040 & 0.230 & 0.260 & 0.150 \\
Llama-3.1-8B-Instruct & 0.170 & 0.030 & 0.050 & 0.340 & 0.300 & 0.320 & 0.670 & 0.050 & 0.090 & 0.530 & 0.860 & 0.660 & 0.430 & 0.310 & 0.280 \\
Qwen2.5-7B-Instruct & 0.390 & 0.050 & 0.090 & 0.220 & 0.720 & 0.340 & 0.430 & 0.420 & 0.420 & 0.630 & 0.330 & 0.440 & 0.420 & 0.380 & 0.320 \\ \midrule
\multicolumn{16}{c}{\textbf{Multimodal Models}} \\ \midrule
CLIP (raw text) & 0.000 & 0.000 & 0.000 & 0.170 & 0.990 & 0.300 & 0.000 & 0.000 & 0.000 & 0.330 & 0.010 & 0.020 & 0.125 & 0.250 & 0.080 \\
CLIP & 0.000 & 0.000 & 0.000 & 0.182 & 0.922 & 0.304 & 0.198 & 0.157 & 0.175 & 0.000 & 0.000 & 0.000 & 0.095 & 0.270 & 0.120 \\
LLaVA-NeXT (P3) & 0.444 & 0.028 & 0.052 & 0.246 & 0.723 & 0.367 & 1.000 & 0.016 & 0.032 & 0.527 & 0.502 & 0.515 & 0.554 & 0.318 & 0.242 \\
Qwen2-VL-7B-Instruct & 0.500 & 0.130 & 0.200 & 0.000 & 0.000 & 0.000 & 0.300 & 0.490 & 0.370 & 0.520 & 0.730 & 0.610 & 0.330 & 0.340 & 0.290 \\
LLaVA-NeXT (P4) & 0.387 & 0.083 & 0.137 & 0.236 & 0.390 & 0.294 & 0.550 & 0.090 & 0.155 & 0.524 & 0.683 & 0.593 & 0.424 & 0.312 & 0.295 \\
Qwen2.5-VL-7B-Instruct & 0.500 & 0.070 & 0.120 & 0.380 & 0.400 & 0.390 & 0.250 & 0.570 & 0.350 & 0.560 & 0.510 & 0.530 & 0.420 & 0.390 & 0.350 \\ \bottomrule
\end{tabular}%
}
\end{table*}

\paragraph{Multimodal Models}
Performance varies significantly among multimodal models, as can be observed from Table \ref{tab:hvvp1results}. Qwen2-VL-7B-Instruct scores lowest (macro F1 = 0.17), despite yielding the best F1 scores in \textit{hero} and \textit{victim} among multimodal models (0.16 and 0.17, respectively). This suggests strong comprehension capabilities of Qwen2-VL-7B-Instruct. CLIP, although slightly better overall, shows no prominent prediction capability for \textit{hero, victim,} and \textit{villain} when examined with raw embedded text, suggesting inadequate contextual embedding for nuanced roles and ending up with a performance equivalent of a majority baseline. Qwen2.5-VL-7B Instruct, however, performs more robustly, with an F1 of 0.28 overall, 0.16 for \textit{hero}, 0.43 for \textit{other}, 0.17 for \textit{victim}, and 0.47 for \textit{villain}. This highlights its stronger multimodal reasoning capabilities.

In further ablation, comparing CLIP-based embeddings using processed OCR vs. raw OCR context reveals that processed OCR inputs leads to slightly better performance, but still fails in categories like \textit{hero} and \textit{victim}, suggesting the OCR-based performance bottleneck for such approaches. LLaVA-NeXT via prompt P3 performs the best overall (F1 = 0.33), and with comparatively decent scores for \textit{hero}, \textit{other} and \textit{villain} role prediction, demonstrating competent instruction-following capability. However, its poor \textit{victim} detection underscores its inherent limitations as a multimodal reasoner.

\subsection{Performance on \hvvdatatwo~(En) test set}

Table \ref{tab:hvvp2Enresults} reports the benchmarking results for the \hvvdatatwo~test set in English. This evaluation, like in \hvvdata, spans a variety of models -- code-mix treated traditional language models fine-tuned on English memes data, and text-only large language models, and multimodal models -- assessed under a zero-shot setting.

\paragraph{Traditional Language Models}
Consistent with performance on \hvvdata~(En), code-mix treated models underperform, suggesting continued limitations in linguistic generalisation. An interesting pattern persists where models fine-tuned on sentiment data exhibit improved detection of \textit{villain} narratives -- likely due to inductive bias from affective language during training. 

While sentiment-based models show some competence in detecting \textit{hero} narratives, muril-large-cased performs slightly worse (F1 = 0.28) than its smaller, fine-tuned counterpart (F1 = 0.31), indicating diminishing returns with model scaling alone. In contrast, deberta-V3-Large emerges as a strong contender, achieving the second-best macro F1 score of 0.28 and the highest Victim prediction score (F1 = 0.32).

\paragraph{Text-only Large Language Models (Zero-shot)}
Among the LLMs evaluated, as can be seen from Table \ref{tab:hvvp2Enresults}, Phi-2 and Mistral-7B-Instruct-v0.1 yield moderate performance on \textit{hero} narrative role detection (0.18 and 0.13, respectively) but struggle significantly with \textit{villain} prediction, scoring just 0.03 and 0.04, respectively. Meanwhile, LlaMA-3-8B-Instruct once again leads in \textit{villain} detection (F1 = 0.66) and achieves an overall macro F1 of 0.28--mirroring its selective strengths observed when evaluated on \hvvdata~(cf. Table \ref{tab:hvvp1results}).

The best overall performance within this group is observed for Qwen-2.5B-Instruct, delivering a macro F1 of 0.32 and the highest \textit{victim} F1 score of 0.44. Across the board, the \textit{hero} category scores remains below the random baseline, reinforcing the challenge of detecting constructive narratives.

\paragraph{Multimodal Models}
Multimodal evaluation reveals modest trends. As in the case of \hvvdata, CLIP with processed text slightly outperforms its raw OCR counterpart overall (0.12 vs 0.08 F1 score), yet remains ineffective in capturing the nuanced context of \textit{hero} narratives. The previously observed trend, where CLIP using raw vs. processed text performs selectively well for the villain and victim categories, further underscores the challenges of achieving generalizability in this task. Among prompt-based variants of LlaVA-Next, the Prompt-4 configuration shows a 4-5\% overall F1 score improvement over Prompt-3, particularly for predicting roles of \textit{hero} and \textit{villain} category prediction. Still, the highest \textit{other} category prediction maxes at just 0.367, and performance for \textit{hero} detection remains poor. However, the models fare better in Villain narrative detection, with F1 scores of 0.51 and 0.59, respectively. This is contrary to \hvvdata~findings, where \textit{other} narratives contributed significantly to overall F1. The more balanced dataset in \hvvdatatwo~may facilitate more equitable assessment of Prompt-4's generalizable predictions. Overall, Macro F1 scores of 0.24 and 0.29 indicate that, while performance exceeds random and majority baselines, inferencing capacity remains limited and class-dependent.

Among all models, Qwen-2.5-VL-7B-Instruct stands out with the best macro-average F1 score of 0.35, backed by strong \textit{victim} (F1 = 0.35) and \textit{villain} (F1 = 0.53) prediction. It is closely followed by Qwen-2-VL-7B-Instruct, which demonstrates the best \textit{hero} detection (F1 = 0.20), \textit{victim} (F1 = 0.37), and \textit{villain} (F1 = 0.61). However, its complete failure to predict the \textit{other} category (F1 = 0.00) pulls its macro F1-score down to 0.30. This reflects a key challenge: high performance in some classes with defined characteristics can be offset by complete failure in others with ambiguous definitions, especially in subtle narrative types. These results also highlight the robust reasoning capabilities of Qwen-2.5-VL-7B-Instruct.

\begin{table*}[t!]
\centering
\caption{Benchmarking results on HVV Phase 2 test set - EnHI (code-mixed).}
\label{tab:hvvp2resultscm}
\resizebox{\textwidth}{!}{%
\begin{tabular}{@{}lccccccccccccccc@{}}
\toprule
 & \multicolumn{3}{c}{\textbf{Hero (252)}} & \multicolumn{3}{c}{\textbf{Other (148)}} & \multicolumn{3}{c}{\textbf{Victim (207)}} & \multicolumn{3}{c}{\textbf{Villain (348)}} & \multicolumn{3}{c}{\textbf{Macro (955)}} \\ \cmidrule(l){2-16} 
\multirow{-2}{*}{} & \textbf{Prec} & \textbf{Rec} & \textbf{F1} & \textbf{Prec} & \textbf{Rec} & \textbf{F1} & \textbf{Prec} & \textbf{Rec} & \textbf{F1} & \textbf{Prec} & \textbf{Rec} & \textbf{F1} & \textbf{Prec} & \textbf{Rec} & \textbf{F1} \\ \midrule
{\color[HTML]{9B9B9B} Majority} & {\color[HTML]{9B9B9B} 0.000} & {\color[HTML]{9B9B9B} 0.000} & {\color[HTML]{9B9B9B} 0.000} & {\color[HTML]{9B9B9B} 0.000} & {\color[HTML]{9B9B9B} 0.000} & {\color[HTML]{9B9B9B} 0.000} & {\color[HTML]{9B9B9B} 0.000} & {\color[HTML]{9B9B9B} 0.000} & {\color[HTML]{9B9B9B} 0.000} & {\color[HTML]{9B9B9B} 0.360} & {\color[HTML]{9B9B9B} 1.000} & {\color[HTML]{9B9B9B} 0.530} & {\color[HTML]{9B9B9B} 0.090} & {\color[HTML]{9B9B9B} 0.250} & {\color[HTML]{9B9B9B} 0.130} \\
{\color[HTML]{9B9B9B} Random} & {\color[HTML]{9B9B9B} 0.280} & {\color[HTML]{9B9B9B} 0.250} & {\color[HTML]{9B9B9B} 0.260} & {\color[HTML]{9B9B9B} 0.150} & {\color[HTML]{9B9B9B} 0.230} & {\color[HTML]{9B9B9B} 0.180} & {\color[HTML]{9B9B9B} 0.230} & {\color[HTML]{9B9B9B} 0.290} & {\color[HTML]{9B9B9B} 0.260} & {\color[HTML]{9B9B9B} 0.400} & {\color[HTML]{9B9B9B} 0.280} & {\color[HTML]{9B9B9B} 0.330} & {\color[HTML]{9B9B9B} 0.260} & {\color[HTML]{9B9B9B} 0.260} & {\color[HTML]{9B9B9B} 0.260} \\ \midrule
\multicolumn{16}{c}{\textbf{Text-Only Models}} \\ \midrule

muril-base-cased & 0 & 0 & 0 & 0.153 & 0.966 & 0.265 & 0.111 & 0.005 & 0.009 & 0.417 & 0.014 & 0.028 & 0.17 & 0.246 & 0.075 \\
bert-base (ml-c) & 0 & 0 & 0 & 0.151 & 0.946 & 0.26 & 0.381 & 0.039 & 0.07 & 0.429 & 0.009 & 0.017 & 0.24 & 0.248 & 0.087 \\
bert-base (ml-uc-sent) & 0 & 0 & 0 & 0.159 & 0.966 & 0.273 & 0 & 0 & 0 & 0.32 & 0.046 & 0.08 & 0.12 & 0.253 & 0.088 \\
deberta-v3-large & 0.545 & 0.024 & 0.046 & 0.157 & 0.973 & 0.271 & 0.313 & 0.024 & 0.045 & 0.5 & 0.017 & 0.033 & 0.379 & 0.26 & 0.099 \\
OGBV-gender-indicbert (cm) & 0.2 & 0.004 & 0.008 & 0.16 & 0.953 & 0.275 & 0.231 & 0.029 & 0.052 & 0.533 & 0.069 & 0.122 & 0.281 & 0.264 & 0.114 \\
bert-base (ml-cm-c-sent) & 0.333 & 0.008 & 0.016 & 0.159 & 0.919 & 0.271 & 0.286 & 0.126 & 0.174 & 0.5 & 0.003 & 0.006 & 0.319 & 0.264 & 0.117 \\
muril-base (c-ft-cm) & 0.364 & 0.016 & 0.03 & 0.16 & 0.912 & 0.272 & 0.286 & 0.029 & 0.053 & 0.392 & 0.089 & 0.145 & 0.3 & 0.262 & 0.125 \\
gk-hinglish-sentiment & 0.175 & 0.028 & 0.048 & 0.162 & 0.865 & 0.273 & 0.184 & 0.068 & 0.099 & 0.438 & 0.06 & 0.106 & 0.24 & 0.255 & 0.131 \\
hinglish11k-sentiment-analysis & 0.273 & 0.139 & 0.184 & 0.152 & 0.75 & 0.252 & 0.176 & 0.058 & 0.087 & 0.667 & 0.052 & 0.096 & 0.317 & 0.25 & 0.155 \\
muril-large (c) & 0.5 & 0.02 & 0.038 & 0.164 & 0.912 & 0.278 & 0.436 & 0.082 & 0.138 & 0.47 & 0.112 & 0.181 & 0.392 & 0.282 & 0.159 \\
hindi-abusive-MuRIL (cm) & 0.284 & 0.083 & 0.129 & 0.172 & 0.77 & 0.281 & 0.266 & 0.101 & 0.147 & 0.37 & 0.147 & 0.21 & 0.273 & 0.275 & 0.192 \\ \midrule
phi-2 & 0.23 & 0.25 & 0.24 & 0.17 & 0.75 & 0.28 & 0.31 & 0.02 & 0.04 & 0.3 & 0.01 & 0.02 & 0.25 & 0.26 & 0.14 \\
Mistral-7B-Instruct-v0.1 & 0.31 & 0.15 & 0.21 & 0.16 & 0.84 & 0.27 & 0.24 & 0.07 & 0.11 & 0 & 0 & 0 & 0.18 & 0.27 & 0.15 \\
Qwen2.5-7B-Instruct & 0.62 & 0.05 & 0.1 & 0.17 & 0.93 & 0.29 & 0.25 & 0.06 & 0.1 & 0.56 & 0.11 & 0.18 & 0.4 & 0.29 & 0.17 \\
Llama-3.1-8B-Instruct & 0.56 & 0.04 & 0.07 & 0.12 & 0.16 & 0.14 & 0.5 & 0 & 0.01 & 0.36 & 0.78 & 0.5 & 0.39 & 0.25 & 0.18 \\ \midrule
\multicolumn{16}{c}{\textbf{Multimodal Models}} \\ \midrule
HVV-CLIP (raw text) & 0 & 0 & 0 & 0.16 & 1 & 0.27 & 0 & 0 & 0 & 1 & 0 & 0.01 & 0.29 & 0.25 & 0.07 \\
CLIP & 0 & 0 & 0 & 0.156 & 0.98 & 0.269 & 0.36 & 0.043 & 0.078 & 0 & 0 & 0 & 0.129 & 0.256 & 0.087 \\
LLaVA-NeXT (P3) & 0.561 & 0.091 & 0.157 & 0.166 & 0.872 & 0.279 & 1 & 0.005 & 0.01 & 0.426 & 0.167 & 0.24 & 0.538 & 0.284 & 0.171 \\
LLaVA-NeXT (P2) & 0.507 & 0.139 & 0.218 & 0.17 & 0.905 & 0.287 & 0.5 & 0.005 & 0.01 & 0.567 & 0.158 & 0.247 & 0.436 & 0.302 & 0.19 \\
LLaVA-NeXT (P1) & 0.314 & 0.456 & 0.372 & 0 & 0 & 0 & 0.182 & 0.01 & 0.018 & 0.375 & 0.621 & 0.468 & 0.218 & 0.272 & 0.215 \\
LLaVA-NeXT (P4) & 0.372 & 0.266 & 0.31 & 0.154 & 0.392 & 0.221 & 0.3 & 0.014 & 0.028 & 0.373 & 0.417 & 0.393 & 0.3 & 0.272 & 0.238 \\
Qwen2-VL-7B-Instruct & 0.39 & 0.41 & 0.4 & 0.14 & 0.02 & 0.04 & 0.3 & 0.41 & 0.35 & 0.4 & 0.45 & 0.42 & 0.31 & 0.32 & 0.3 \\ 
Qwen2.5-VL-7B-Instruct & 0.36 & 0.17 & 0.23 & 0.18 & 0.3 & 0.22 & 0.31 & 0.46 & 0.37 & 0.41 & 0.32 & 0.36 & 0.31 & 0.31 & 0.3 \\\bottomrule
\end{tabular}%
}
\end{table*}

\subsection{Performance on the \hvvdatatwo~(En-Hi) test set}

The majority of the models, like the multimodal model (CLIP), smaller models like muril-base, and bert-base-multilingual-uncased-sentiment, seem to have high recall rates for the \textit{other} category but have the lowest macro F1 scores. This suggests that bigger models have been beneficial for the code-mixed setting, and smaller models, even with multi-lingual/code-mixed pre-training, do not make the cut. On the other hand, the majority representation leads to the lowest performance of 0.75 as the recall for the \textit{other} category.

For the \textit{hero} category, no smaller model can make reasonable recognition. Both Google-muril-large and Deberta-large have good precision, but barely have a recall rate. Both CNERG-cm-abusive-muril and 11k-sentiment-analysis have comparatively better recall rates, whereas LlaVa-Next prompts 1 and 2 have better F1 scores for the hero category, the former has the highest F1 score of 0.37. This further consolidates the improvement induced by the code-mixed pretraining involved in models, with LlaVa-Next having overall better multi-lingual and reasoning capabilities to recognise the positive portrayals within code-mixed memes.

Interestingly, for the \textit{victim} category, some (read finite/not-encouraging) of the victim cases identified are correctly labelled by LlaVa-Next-based models; they perform poorly on the recall and F1 scores, suggesting difficulty in detecting the subtle victimisations. On the other hand, models that are not only pre-trained using code-mixed data but are also fed with abusive/sentiment/hate-speech data, perform better than those fed with only code-mixed data (muril-base (c-ft-cm), bert-based-ml-case, deberta-v3-large, OGBV-gender-indicbert (cm), muril-base-cased). bert-base (ml-cm-c-sent), which is purely a multilingual model \cite{khanuja-etal-2020-gluecos}, focusing on code-mix sentiment analysis, performes exceptionally well for victim identification. However, it does not perform as effectively for villain identification -- a very interesting observation. Similar to its performance in victim identification, sentiment analysis-based models shows mediocre performance overall, slightly outperforming Microsoft's V3 large monolingual model and several other smaller models, especially those trained solely on code-mix data. Notably, models specifically trained on some form of hate speech-related data, such as OGBV-gender-indicbert (cm) and hindi-abusive-MuRIL (cm), demonstrate slightly better F1 scores compared to purely sentiment-based models. Additionally, the muril-large (c) model performs quite well for detecting villains. Finally, LlaVA-Next-based models exhibit the best performance among the models tested, with prompts \#2 and \#3 not performing as effectively as prompt \#1. This is the case even without explicitly emphasising the \textit{other} category or its definitions. Although these models perform poorly in detecting the \textit{other} category, they provide satisfactory results for the remaining categories, suggesting a potential area for improvement by combining basic crisp instructions with task descriptions and clear definitions of the categories involved.



\begin{figure*}
    \centering
    \includegraphics[width=1\linewidth]{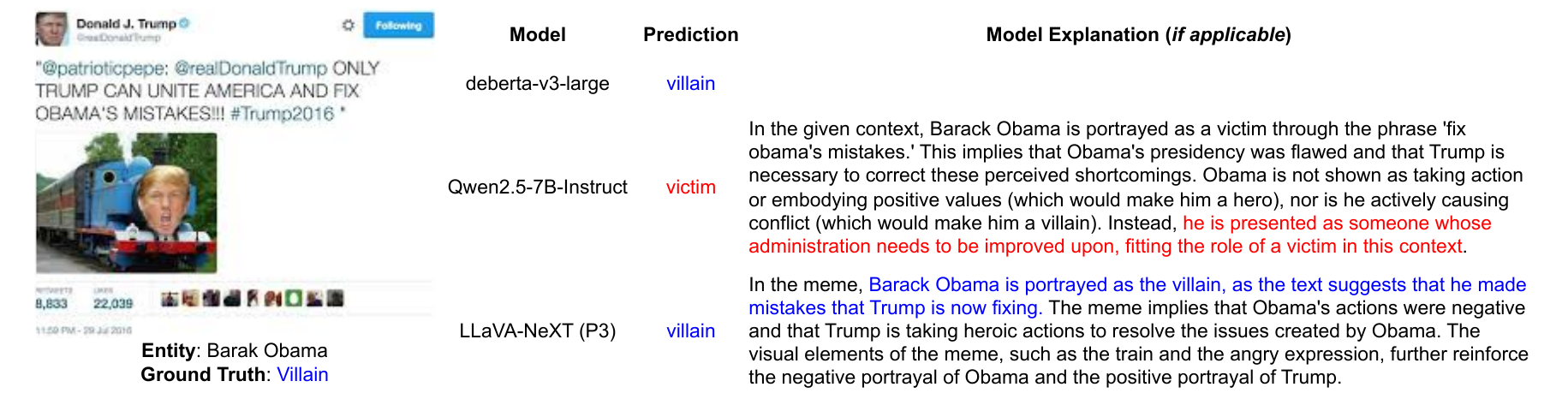}
    \caption{Qualitative examples for HVV-P1-EN, top 3 models from each category.}
    \label{fig:qa_hvv1_en}
\end{figure*}

\section{Analysis}
This section provides a multi-faceted analysis of model behaviour across varied narrative role detection scenarios in memes. We examine model interpretability, representational robustness, and error tendencies through three complementary lenses: (i) Qualitative analysis, which inspects model reasoning on representative examples across linguistic and multimodal contexts; (ii) Diagnostic embedding analysis, which reveals latent semantic structures and cultural-linguistic patterns captured by model embeddings; and (iii) Error analysis, which identifies systematic misclassifications and challenges in detecting nuanced roles through confusion matrix patterns. Together, these analyses offer a comprehensive understanding of the strengths, limitations, and cultural sensitivities of state-of-the-art models in handling complex and ambiguous meme content.

\paragraph{Qualitative Analysis} We present a qualitative comparison of model outputs on one representative example from each of the three test sets: \hvvdata~(En), \hvvdatatwo~(En), and \hvvdatatwo~(EnHi). The analysis highlights the interpretive strengths and limitations of top-performing models across varied linguistic and multimodal contexts. We present the demonstrative figure for \hvvdata~(En) here, and include the figures for \hvvdatatwo~(En) and (EnHi) in Appendix A in supplementary.

On the \hvvdata~(En) example (cf. Figure \ref{fig:qa_hvv1_en}), deberta-v3-large accurately identifies the intended narrative role, demonstrating strong alignment with semantic cues. In contrast, Qwen2.5-7B-Instruct, which lacks multimodal capabilities, fails to interpret the meme correctly. LLaVA-NeXT (P3), however, benefits from multimodal cue integration, significantly improving its interpretive accuracy.

For the \hvvdatatwo~(En) sample (cf. Appendix A in supplementary, Figure 8, muril-base (fine-tuned on code-mixed data) struggles to transfer relevant knowledge and produces an incorrect prediction. Qwen2.5-7B-Instruct arrives at the correct answer, but more by chance than through grounded reasoning. By contrast, Qwen2.5-VL-7B-Instruct leverages visual inputs effectively, generating reasoning that is more coherent and semantically aligned with the meme’s intended message.

Finally, in the \hvvdatatwo~(EnHi) case (cf. Appendix A in supplementary, Figure 9), hindi-abusive-MuRIL misclassifies the meme under the \textit{other} category, failing to capture nuanced semantic cues. Llama-3.1-8B-Instruct performs a reasonable translation but does not provide a conclusive or contextually grounded response, highlighting its limitations in ambiguous, code-mixed scenarios. Qwen2.5-VL-7B-Instruct excels by correctly interpreting the idiomatic Hinglish expression (e.g., ``karne wale bohot hai, khatam karne wala ek hi tha'') even without relying on visual cues, demonstrating its robustness in handling multilingual, figurative content.

\begin{figure}[t!]
    \centering
    \includegraphics[width=\linewidth]{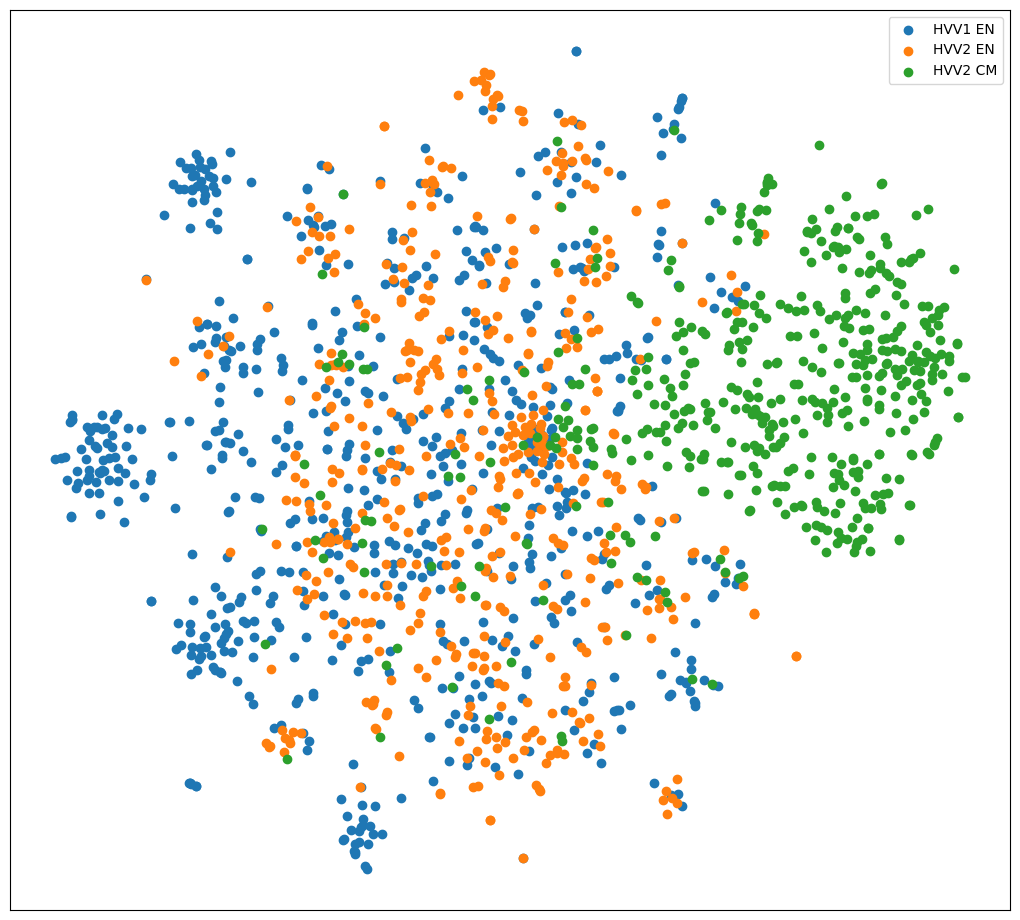}
    \caption{t-SNE plot visualising the data points from the three test sets: \hvvdata~(En) -- \textit{blue}, \hvvdatatwo~(En) -- \textit{orange}, and \hvvdatatwo~(HiEn) or code-mixed (CM) -- \textit{green}.}
    \label{fig:tsne}
\end{figure}

\begin{figure*}[t!]

    \centering
    \includegraphics[width=1\linewidth]{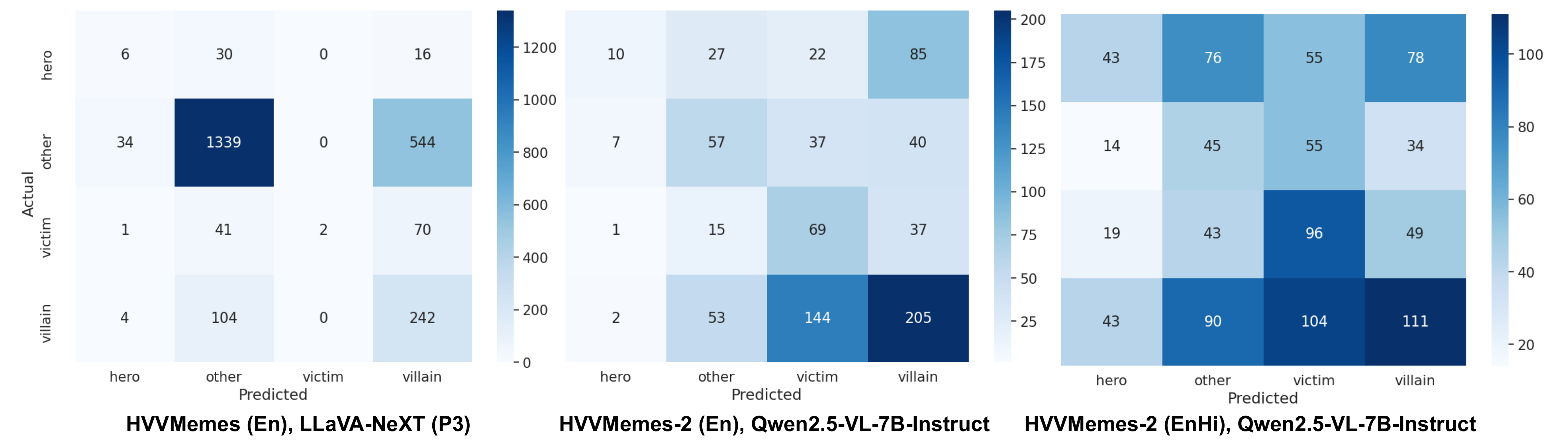}
    \caption{Confusion matrices for the top-performing multimodal models across different test sets. \hvvdata~(En) -- \textit{left}, \hvvdatatwo~(En) -- \textit{middle}, and \hvvdatatwo~(EnHi) -- \textit{right}.}
    \label{fig:small_confmatrices}
\end{figure*}


\paragraph{Diagnostic Analysis} To investigate the semantic and cultural nuances encoded within the OCR-extracted textual content of memes, we first perform a diagnostic assessment of the semantic content via t-SNE-based embedding visualisation (cf. Figure \ref{fig:tsne}) across three distinct test sets: \hvvdata~(En), \hvvdatatwo~(En), and \hvvdatatwo~(EnHi). The resulting plot reveals insightful patterns that highlight the differentiation and overlap among these datasets.

Notably, the \hvvdatatwo~(EnHi) test set, which primarily consists of code-mixed text (e.g., Hinglish), forms a compact and well-defined cluster distinctly separated from the other two sets. This tight clustering suggests a strong linguistic and cultural homogeneity within the code-mixed memes, reflecting a specific regional context that is consistently encoded by the model's embeddings.

In contrast, the \hvvdata~(En) set displays multiple smaller, dense clusters dispersed across the embedding space. This distribution indicates the presence of diverse semantic subgroups or themes within the English meme content, likely driven by a wide range of topics or discourse reflective of US-centric social media. Such fragmentation suggests that the model captures fine-grained thematic distinctions in this dataset.

The \hvvdatatwo~(En) dataset occupies an intermediate and more diffuse region in the semantic space, overlapping significantly with \hvvdata~(En) points. This positioning implies that \hvvdata~(En) memes reference a broader and more global set of themes, potentially including protected or sensitive categories, thereby bridging the culturally specific clusters of \hvvdata~(En) and \hvvdatatwo~(EnHi).

The two extremes along the horizontal axis of the t-SNE plot underscore the cultural and linguistic spectrum present in the datasets: the clustered \hvvdatatwo~(EnHi), on the one end, and the multi-clustered \hvvdata~(En), on the other hand, with \hvvdatatwo~(En) acting as a semantic intermediary. This gradient reflects the embeddings' capacity to encode not only language but also culturally nuanced textual semantics in meme content.

Overall, the embedding analysis demonstrates the model's effectiveness in capturing distinct cultural-linguistic characteristics across datasets, providing a valuable lens for understanding how meme text representations vary according to language style and regional discourse.


\paragraph{Error Analysis}
The confusion matrix analysis reveals that narrative role classification in memes remains a challenging task, particularly due to cultural nuance and role ambiguity. Categories like \textit{other} frequently absorb misclassifications, especially from \textit{villain} and \textit{victim}, underscoring the difficulty in modelling subtle framing cues. While models like Qwen2.5-VL-7B-Instruct exhibit improved semantic generalisation, especially in multimodal and code-mixed scenarios, others tend to either overpredict stereotyped roles or resort to conservative classifications. These trends indicate that while multimodal LLMs offer promise, they still face trade-offs between robustness, safety alignment, and multilingual capability. We demonstrate the confusion matrices for the top-performing multimodal models across the test sets in Figure \ref{fig:small_confmatrices}.\footnote{See Appendix A in supplementary for further details on error analysis.}
\section{Conclusion}
In this work, we introduced and benchmarked the dataset variants for the complex task of narrative role recognition in memes, focusing on four roles: \textit{hero, villain, victim, and other}. Through a comprehensive evaluation across realistic and code-mixed settings, we observed that traditional models and sentiment-tuned variants showed partial success, particularly for villain and other roles, while struggling with nuanced and constructive portrayals like hero and victim. Larger instruction-tuned and multimodal models, particularly those like Qwen2.5-VL-7B-Instruct and LLaVA-NeXT (P4), demonstrated promising improvements--yet their performance remained highly class-dependent and sensitive to prompt design. Lexical and structural analysis further underscored the cultural and semantic variance across datasets, pointing to the need for better multimodal reasoning, especially in code-mixed contexts.

Overall, while prompt-based hybrid strategies and multimodal inputs enhanced robustness, victim detection emerged as a persistent challenge, calling for deeper contextual modelling and culturally aware fine-tuning. Future directions include more explicit representation learning for subtle narrative cues, incorporation of socio-pragmatic features, and the development of role-adaptive training objectives to improve generalisability across diverse meme styles and languages.

\appendices

\begin{figure*}[t!]
\centering
\subfloat[{\centering \hvvdata~(En)}\label{fig:hvv1enwc}]{
\includegraphics[width=0.30\textwidth]{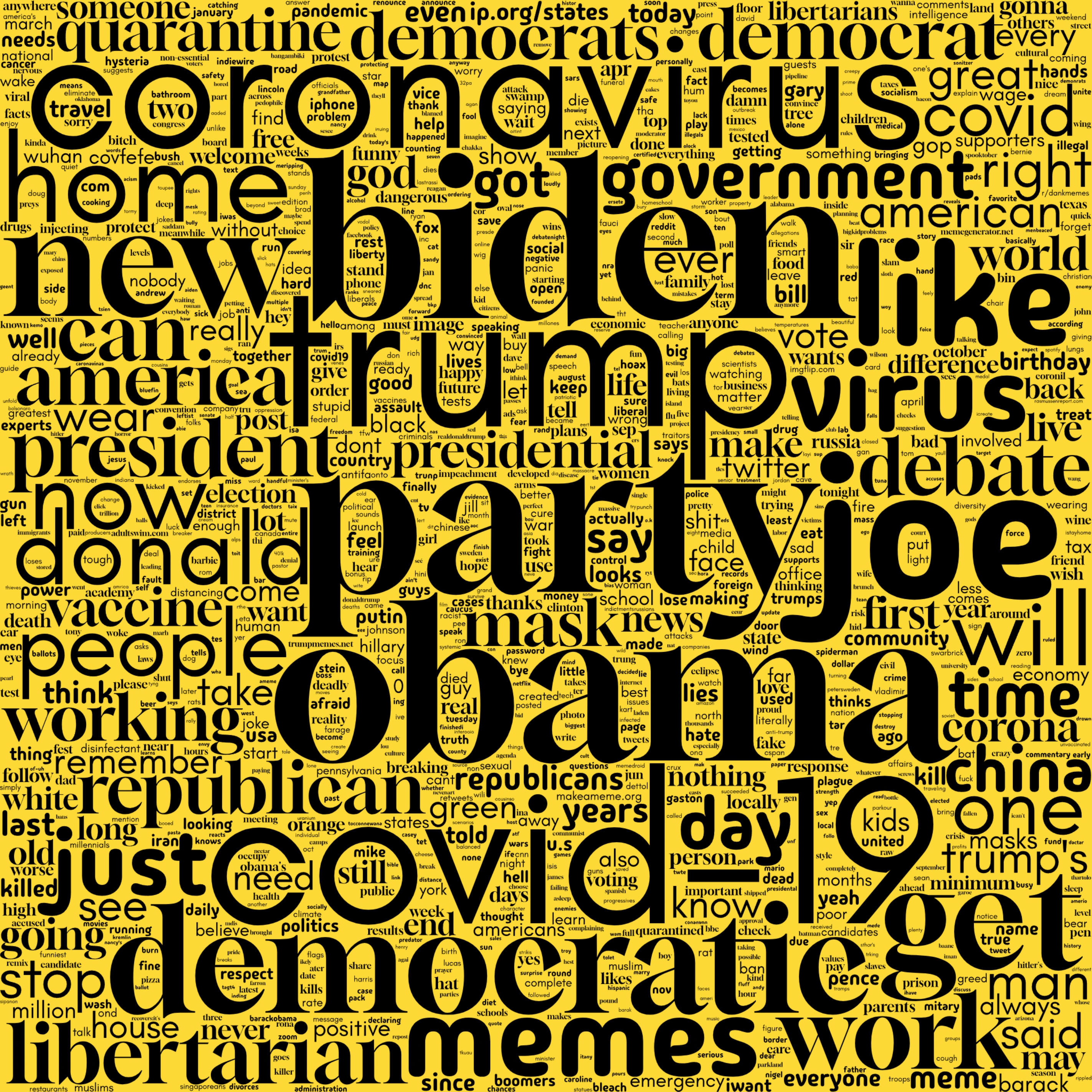}}\hspace{0.1em}
\subfloat[{\hvvdatatwo~(En)}\label{fig:hvv2enwc}]{
\includegraphics[width=0.30\textwidth]{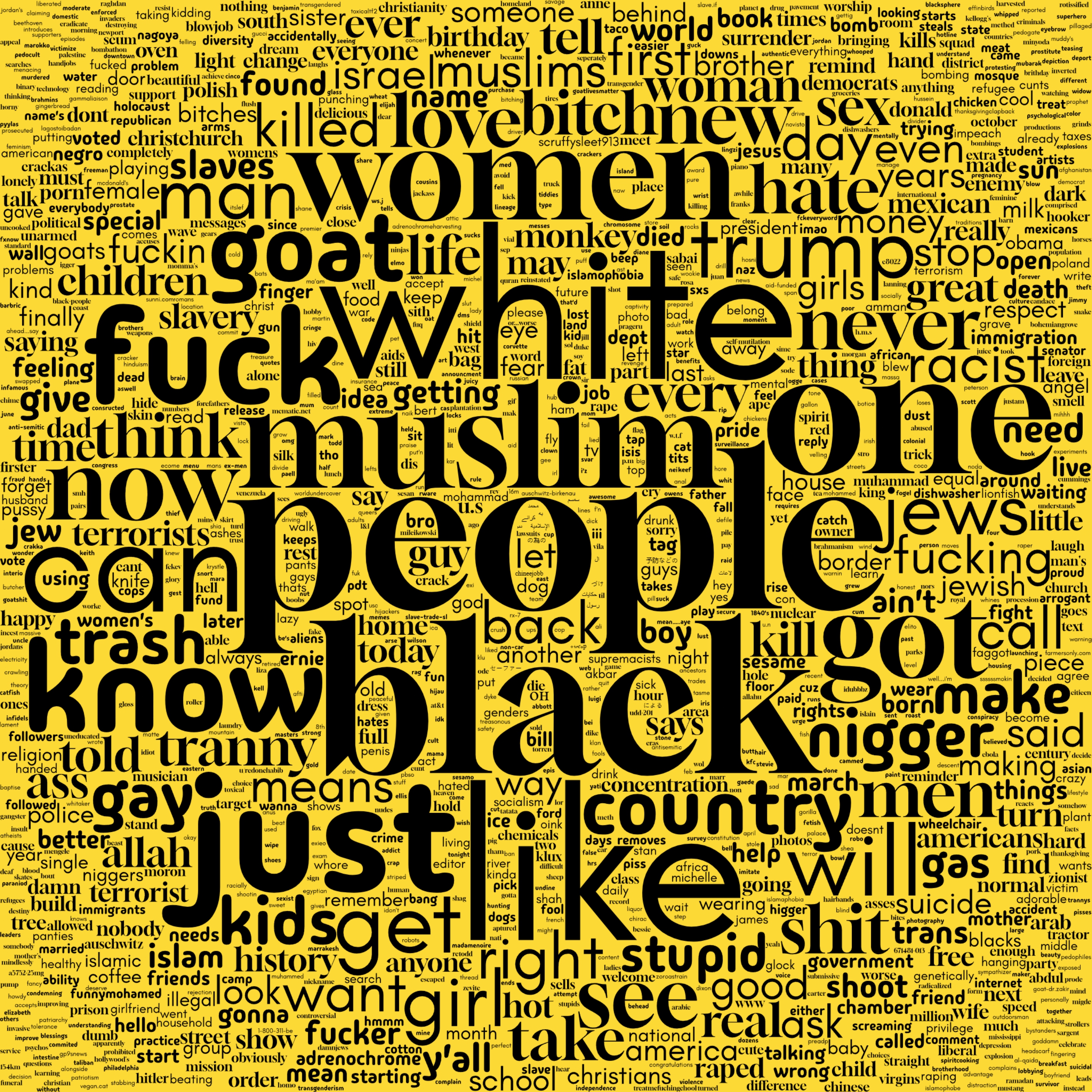}}\hspace{0.1em}
\subfloat[{\centering \hvvdatatwo~(EnHi)}\label{fig:hvv2cmwc}]{
\includegraphics[width=0.30\textwidth]{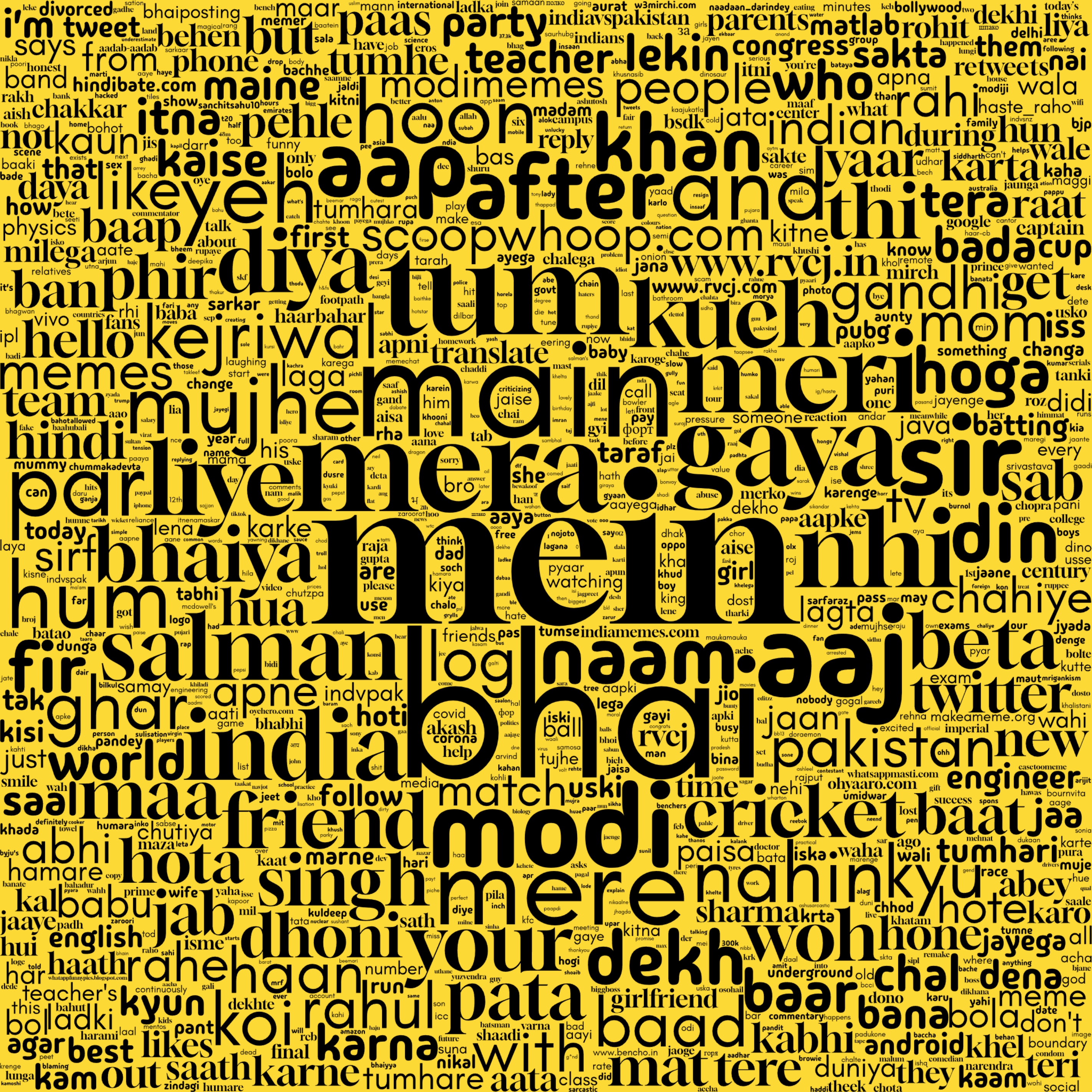}} 
\caption{Word cloud representations for \hvvdata\ (En) -- \textit{left}, \hvvdatatwo~(En) -- \textit{middle}, and \hvvdatatwo~(EnHi) -- \textit{right}.}
\label{fig:wordclouds}
\end{figure*}

\section{}
This section provides supplementary insights supporting the main study. First, we present a lexical analysis of three test sets using word cloud visualisations, revealing thematic and linguistic contrasts between real-world and synthetic meme datasets across English and code-mixed Hindi-English (Hinglish) content. We then detail the suite of baseline models used, spanning multilingual, code-mixed, sentiment, vision-language, and instruction-tuned architectures. Comprehensive error analyses, including confusion matrices, highlight model-specific trends and role classification challenges in nuanced meme narratives. Finally, we include qualitative examples demonstrating model outputs across datasets and enumerate the prompt variants used for consistent evaluation. 

\section*{Lexical Analysis via Word Clouds}
\label{app:appstart}
\new{The three test sets exhibit notable distinctions in lexical characteristics, as observed through their respective word clouds (see Figure \ref{fig:wordclouds}). The first and third test sets, \hvvdata~(En) and \hvvdatatwo~(EnHi) (Figures \ref{fig:hvv1enwc} and \ref{fig:hvv2cmwc}, respectively), both comprising real memes, present subtle and contextually rich lexicon reflective of real-world themes. \hvvdata~(En) includes English-language memes centred on U.S. politics and COVID-19, with commonly occurring terms such as `party', `Joe', `Obama', `COVID-19', `coronavirus', and `democratic'. The third test set, which contains code-mixed Hindi-English memes, features culturally specific terms like `Bhai', `Modi', `Sir', `Khan', `cricket', `Kejriwal', etc., indicative of Indian political and cricket-related themes. In contrast, the second test set, \hvvdatatwo~(En) (see Figure \ref{fig:hvv2enwc}), derived from the synthetically curated Hateful Memes dataset \cite{kiela2020hateful}, displays a lexicon heavily loaded with explicit hate speech terminology. Prominent terms include `f*ck', `white', `Muslim', `nigger', `women', and `hate', which are strongly associated with targeted hate and discriminatory language against protected groups.}

\new{These task differences underline a key challenge: \hvvdata~(En) and \hvvdatatwo~(EnHi), being rooted in real meme content, rely on nuanced and context-driven language that poses challenges to assigning the underlying narrative roles. Conversely, the second dataset's overtly hateful lexicon makes it more straightforward for humans to identify patterns of hate, potentially reducing the complexity of the detection task.}

\section*{Details of comparative baselines}
\paragraph{Multilingual and Code-Mixed Models}
We include a suite of pretrained language models known for their competence in multilingual and code-mixed language understanding. \textbf{MuRIL-base-cased} \cite{khanuja2021muril}, trained on 17 Indian languages, provides a strong foundation for Hindi and Hinglish processing while preserving casing information. We also use \textbf{bert-base} \cite{devlin-etal-2019-bert} in multiple variants -- \textit{ml-cm-c-sent} (fine-tuned for code-mixed sentiment with casing) \cite{khanuja-etal-2020-gluecos}, \textit{ml-c} (multilingual cased) \cite{DBLP:journals/corr/abs-1810-04805}, and \textit{ml-uc-sent} (applied to multilingual uncased sentiment tasks)\footnote{\url{https://huggingface.co/google-bert/bert-base-multilingual-cased}}. The \textbf{muril-base (c-ft-cm)} \footnote{\url{https://huggingface.co/IIIT-L/muril-base-cased-finetuned-code-mixed-DS}} and \textbf{muril-large (c)} \cite{khanuja2021muril} variants further explore the contextual fine-tuning and scale benefits of MuRIL for code-mixed classification tasks.

\paragraph{Hinglish and Social Media Sentiment Models}
Recognising the linguistic peculiarities of social media content, particularly in transliterated and code-switched Hinglish, we incorporate several tailored models. \textbf{hinglish11k-sentiment-analysis}\footnote{\url{https://huggingface.co/yj2773/hinglish11k-sentiment-analysis}} and \textbf{gk-hinglish-sentiment} \cite{khanuja-etal-2020-gluecos} are fine-tuned BERT variants trained on Hinglish sentiment datasets, optimized to handle colloquial expressions. To address socio-linguistic toxicity and bias, \textbf{OGBV-gender-indicbert (cm)}\footnote{\url{https://huggingface.co/Maha/OGBV-gender-indicbert-ta-hasoc21_codemix}} targets gender bias in online gender-based violence contexts, while \textbf{hindi-abusive-MuRIL (cm)} \cite{das2022data} focuses on abusive language detection in Hindi-English code-mixed data.


\paragraph{General-Purpose Instruction-Tuned Models}
We benchmark modern LLMs fine-tuned with instruction-following capabilities. These include \textbf{Llama-3.1-8B-Instruct} \cite{grattafiori2024llama3herdmodels}, a robust open-weight model with 8B parameters, and \textbf{phi-2}\footnote{\url{https://huggingface.co/microsoft/phi-2}}, a compact transformer trained on synthetic and curated corpora for efficiency. \textbf{Mistral-7B-Instruct-v0.1} \cite{jiang2023mistral7b} and \textbf{Qwen2.5-7B-Instruct} \cite{qwen2.5} offer strong reasoning capabilities, while \textbf{deberta-v3-large} \cite{he2021debertav3} provides a competitive transformer backbone with disentangled attention and improved generalisation for classification tasks.

\paragraph{Vision-Language and Multimodal Models}
To explore the multimodal dimension, we include several recent vision-language models. \textbf{Qwen2-VL-7B-Instruct} \cite{Qwen2VL} and \textbf{Qwen2.5-VL-7B-Instruct} \cite{qwen2.5-VL} combine visual input with powerful language reasoning. We also evaluate \textbf{LLaVA-NeXT} \cite{liu2024llavanext} with two empirically designed prompt configurations -- \textit{P3}, which uses detailed descriptions of roles, and \textit{P4}, a hybrid variant combining role information in a simple manner. Furthermore, we include \textbf{CLIP} \cite{radford2021learningtransferablevisualmodels}, a model pretrained to align images and text through contrastive learning, used here in both its full multimodal form and its isolated \textit{text encoder} to measure language-only performance.

Together, these baselines allow us to assess a range of capabilities -- contextual reasoning, sentiment classification, code-mixed understanding, and multimodal grounding -- across varied architectures and training paradigms.

\section*{Detailed Error Analysis}

\begin{figure*}[t!]

    \centering
    \includegraphics[width=1\linewidth]{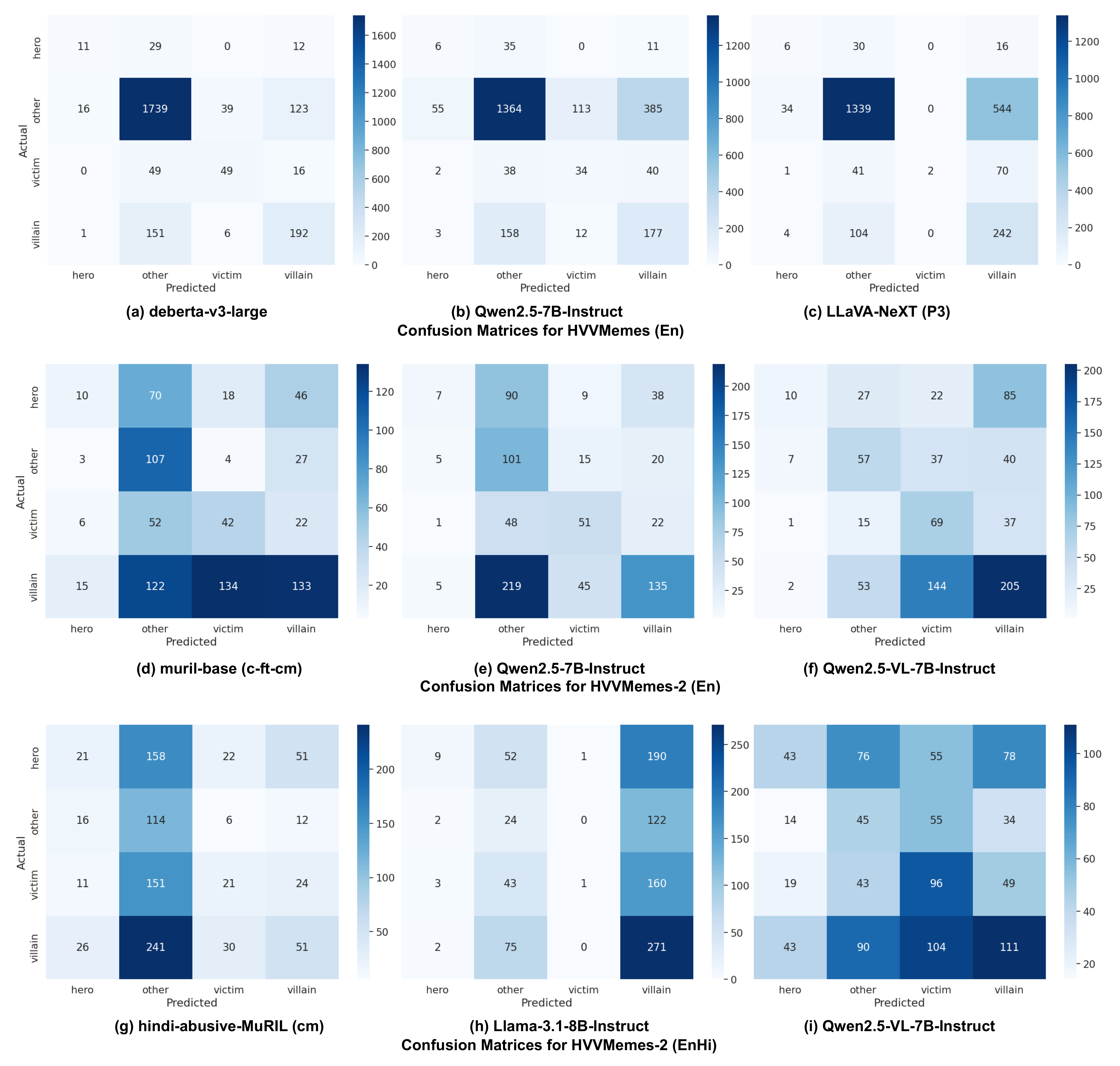}
    \caption{Confusion matrices across top three performant models across different baseline categories and test sets. \hvvdata~(En) -- \textit{top}, \hvvdatatwo~(En) -- \textit{middle}, and \hvvdatatwo~(EnHi) -- \textit{bottom}.}
    \label{fig:confmatrices}
\end{figure*}
\paragraph{Error Analysis} The confusion matrices (cf. Figure \ref{fig:confmatrices}) across different models and test sets reveal insightful patterns about the challenges in classifying nuanced narrative roles in memes.

For the \hvvdata~(En) dataset, a prevalent pattern is the high degree of confusion involving the \textit{other} category. Notably, there is significant misclassification between the \textit{other} and \textit{villain} categories across various model types. This trend aligns with the overall category-wise prediction accuracy, which closely mirrors the distribution of the training data. Particularly, the \textit{victim} role shows markedly poorer prediction performance for the multimodal model LLAVA-NeXT (P3), underscoring the complexities that this category introduces for multimodal models. Such skewed category-specific performance highlights the persistent ambiguities inherent in theme-based memes, where subtle narrative framing makes clear distinctions difficult.

In the case of \hvvdatatwo~(En), the Qwen2.5-VL-7B-Instruct model demonstrates relative gains in predicting the \textit{victim} and \textit{villain} categories compared to other models. However, it also shows a tendency to confuse instances with the \textit{other} category, indicating difficulty in handling ambiguous cases. Models like muril-base (c-ft-cm) and Qwen2.5-7B-Instruct exhibit a conservative approach by frequently misclassifying categories as \textit{other}, which may also point to hallucinations introduced by current multimodal large language models. For the \hvvdatatwo~(En) dataset, models show a strong semantic affinity for the \textit{villain} category, reflecting how explicit societal stereotypes and biases are embedded in the textual semantics, which the models pick up effectively.

Finally, the \hvvdatatwo~(EnHi) (Hinglish) test set presents a nuanced landscape. Multimodal LLMs like hindi-abusive-MuRIL (cm) tend to confuse multiple categories with \textit{Other}, indicating challenges with ambiguity in code-mixed data. Conversely, Llama-3.1-8B-Instruct shows a skewed tendency to overpredict the \textit{villain} category while exhibiting notably poor performance on \textit{hero} and \textit{victim} roles. This may be attributed to the model’s guardrails, designed to mitigate unsafe content or limited support for multiple languages. On the other hand, Qwen2.5-VL-7B-Instruct demonstrates a more balanced semantic generalisation, with some confusion between \textit{hero} and \textit{other} but relatively stronger predictions for \textit{victim} and \textit{villain}. This reflects the model’s enhanced reasoning ability in complex multimodal and code-mixed scenarios, such as Hinglish memes.

Overall, the confusion matrix analyses underscore persistent challenges in modelling subtle and culturally nuanced meme content, particularly in distinguishing less-represented or more ambiguous narrative roles. Multimodal models, while promising, still face hurdles in balancing safety, multilingual support, and nuanced semantic understanding.

\section*{Additional demonstrations from Qualitative Analysis}
\begin{figure*}
    \centering
    \includegraphics[width=1\linewidth]{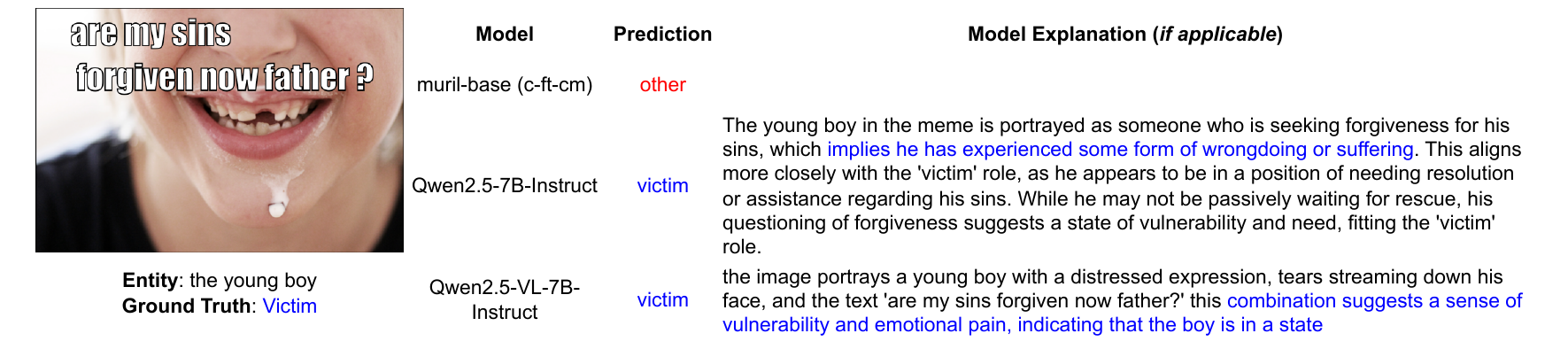}
    \caption{Qualitative examples for HVV-P2-EN, top three models from each category.}
    \label{fig:qa_hvv2_en}
\end{figure*}

\begin{figure*}
    \centering
    \includegraphics[width=1\linewidth]{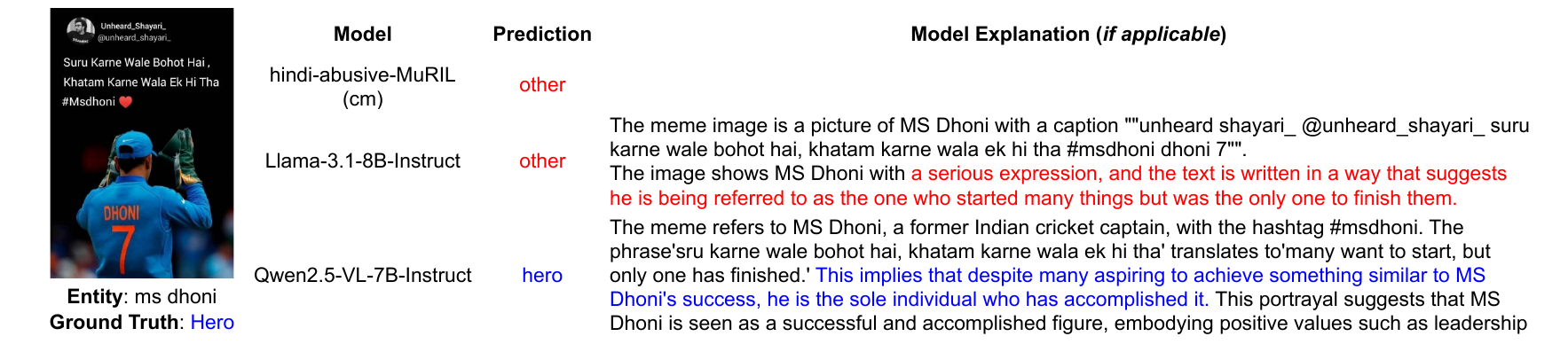}
    \caption{Qualitative examples for HVV-P2-CM, top three models from each category.}
    \label{fig:qa_hvv2_cm}
\end{figure*}

We have included the demonstrative examples towards qualitative analysis as part of this supplementary material in Figure \ref{fig:qa_hvv2_en} and \ref{fig:qa_hvv2_cm}.

\section*{Prompts variants utilised in this work}
\label{app:promptappendix}

\begin{tcolorbox}[
  colback=gray!10,         
  colframe=black,          
  boxrule=0.5pt,           
  arc=3pt,                 
  boxsep=4pt,              
  left=6pt, right=6pt,     
  top=6pt, bottom=6pt      
]
\textbf{\underline{Instruction Text}:}\\
Put this output into a JSON with keys ``answer'', having any ``ONE'' of the given options as a value, and ``explanation'', having justification of the answer choice. Please analyze the image and embedded text carefully, considering visual elements, context, and any implied meanings, before making your decision.
\end{tcolorbox}

\begin{tcolorbox}[
  colback=gray!10,         
  colframe=black,          
  boxrule=0.5pt,           
  arc=3pt,                 
  boxsep=4pt,              
  left=6pt, right=6pt,     
  top=6pt, bottom=6pt      
]
\textbf{\underline{Prompt 1}:} \textit{A basic prompt that was used (without optimisation).}\\\\
You are a critical Question/Answer assistant.
Given the context: \textcolor{blue}{\texttt{\textbf{\{ocr\}}}}
I will provide you with a meme image, and you need to decide and justify, after critically assessing, which of the four narrative roles provided best represents the portrayal of the entity: \textcolor{blue}{\texttt{\textbf{\{entity\}}}}, within the given meme.
Options are: ``hero, villain, victim, and other''. \\\\          \textcolor{blue}{\texttt{\textbf{<instruction text>}}}
\end{tcolorbox}

\begin{tcolorbox}[breakable,
  colback=gray!10,         
  colframe=black,          
  boxrule=0.5pt,           
  arc=3pt,                 
  boxsep=4pt,              
  left=6pt, right=6pt,     
  top=6pt, bottom=6pt      
]
\textbf{\underline{Prompt 2}:} \textit{An explicit statement emphasising the `other' category was added to the prompt.}\\\\
You are a critical Question/Answer assistant.
Given the context: \textcolor{blue}{\texttt{\textbf{\{ocr\}}}}
I will provide you with a meme image, and you need to decide and justify, after critically assessing, which of the four narrative roles provided best represents the portrayal of the entity: \textcolor{blue}{\texttt{\textbf{\{entity\}}}}, within the given meme.
Options are: ``hero, villain, victim, and other''.
If the entity is not depicted as a hero, villain, or victim, you may categorise it under the ``other'' label.\\\\
\textcolor{blue}{\texttt{\textbf{<instruction text>}}}
\end{tcolorbox}

\begin{tcolorbox}[breakable,
  colback=gray!10,         
  colframe=black,          
  boxrule=0.5pt,           
  arc=3pt,                 
  boxsep=4pt,              
  left=6pt, right=6pt,     
  top=6pt, bottom=6pt      
]
\textbf{\underline{Prompt 3}:} \textit{GPT4 optimised prompt, with a definition for all the categories and recommended formatting instruction.}\\\\
Given a meme image with \textcolor{blue}{\texttt{\textbf{<embedded text>}}} extracted via OCR, analyze and categorize the \textcolor{blue}{\texttt{\textbf{<queried entity>}}} within the meme as one of the following labels: `hero', `villain', `victim', or `other'. Assign only one category based on how the entity is depicted, including visual cues, text context, and any implied meanings. Return your response in the following \textcolor{blue}{\texttt{\textbf{<JSON format>}}} with two items:\\\\
\textcolor{blue}{\texttt{\textbf{<JSON format>}}}:\\
first item - key:``answer'', value:``label'' and second item - key:``explanation'', value:``Justification for label choice based on meme analysis, visual elements, context, and implied meanings.''\\
\\
Carefully evaluate the portrayal of the entity considering positive, negative, or neutral framing in the meme context, provided here under the \textcolor{blue}{\texttt{\textbf{<INFORMATION GIVEN>}}} tag. For additional context, also given below are the \textcolor{blue}{\texttt{\textbf{<role definition>}}} of each of the role-label choices.\\\\
\textcolor{blue}{\texttt{\textbf{<role definition>}}}:\\
Hero: The character who takes courageous actions to resolve challenges, embodying positive values and driving progress or resolution.\\
Villain: The character who opposes or creates conflict, often driven by negative motives or obstructing progress.\\
Victim: The character who suffers harm or injustice, often needing rescue, protection, or resolution of their predicament.\\
Other: Any character or entity that does not distinctly fit into the roles of hero, villain, or victim; typically includes neutral, peripheral, or multifaceted figures contributing context, balance, or alternative perspectives.\\\\
\textcolor{blue}{\texttt{\textbf{<INFORMATION GIVEN>}}}:\\
\textcolor{red}{\texttt{\textbf{<embedded text>}}}: \{ocr\}\\
\textcolor{red}{\texttt{\textbf{<queried entity>}}}: \{entity\}\\
The answer in the required JSON format is:
\end{tcolorbox}

\begin{tcolorbox}[breakable,
  colback=gray!10,         
  colframe=black,          
  boxrule=0.5pt,           
  arc=3pt,                 
  boxsep=4pt,              
  left=6pt, right=6pt,     
  top=6pt, bottom=6pt      
]
\textbf{\underline{Prompt 4}:} \textit{A hybrid prompt specifies definitions for narrative-role labels and a custom system instruction.}\\\\
You are a critical Question/Answer assistant.
Given the context: \textcolor{blue}{\texttt{\textbf{\{ocr\}}}}
I will provide you with a meme image, and you need to decide and justify, after critically assessing, which of the four narrative roles provided best represents the portrayal of the entity: \textcolor{blue}{\texttt{\textbf{\{entity\}}}}, within the given meme.
Options are: "hero, villain, victim, and other". \\\\
\textcolor{blue}{\texttt{\textbf{<role definition>}}}:\\
Hero: The character who takes courageous actions to resolve challenges, embodying positive values and driving progress or resolution.\\
Villain: The character who opposes or creates conflict, often driven by negative motives or obstructing progress.\\
Victim: The character who suffers harm or injustice, often needing rescue, protection, or resolution of their predicament.\\
Other: Any character or entity that does not distinctly fit into the roles of hero, villain, or victim; typically includes neutral, peripheral, or multifaceted figures contributing context, balance, or alternative perspectives.\\\\
\textcolor{blue}{\texttt{\textbf{\{instruction text\}}}}
\end{tcolorbox}

\bibliographystyle{IEEEtran}
\bibliography{hvv1,custom}

\begin{thebibliography}{10}
\providecommand{\url}[1]{#1}
\csname url@samestyle\endcsname
\providecommand{\newblock}{\relax}
\providecommand{\bibinfo}[2]{#2}
\providecommand{\BIBentrySTDinterwordspacing}{\spaceskip=0pt\relax}
\providecommand{\BIBentryALTinterwordstretchfactor}{4}
\providecommand{\BIBentryALTinterwordspacing}{\spaceskip=\fontdimen2\font plus
\BIBentryALTinterwordstretchfactor\fontdimen3\font minus \fontdimen4\font\relax}
\providecommand{\BIBforeignlanguage}[2]{{%
\expandafter\ifx\csname l@#1\endcsname\relax
\typeout{** WARNING: IEEEtran.bst: No hyphenation pattern has been}%
\typeout{** loaded for the language `#1'. Using the pattern for}%
\typeout{** the default language instead.}%
\else
\language=\csname l@#1\endcsname
\fi
#2}}
\providecommand{\BIBdecl}{\relax}
\BIBdecl

\bibitem{MacAvaney2019Hate}
\BIBentryALTinterwordspacing
S.~MacAvaney, H.-R. Yao, E.~Yang, K.~Russell, N.~Goharian, and O.~Frieder, ``Hate speech detection: Challenges and solutions,'' \emph{PLOS ONE}, vol.~14, no.~8, pp. 1--16, 08 2019. [Online]. Available: \url{https://doi.org/10.1371/journal.pone.0221152}
\BIBentrySTDinterwordspacing

\bibitem{kiela2020hateful}
D.~Kiela, H.~Firooz, A.~Mohan \emph{et~al.}, ``The hateful memes challenge: Detecting hate speech in multimodal memes,'' in \emph{Proceedings of the 34th International Conference on Neural Information Processing Systems}, ser. NeurIPS~'20, vol.~33, 2020.

\bibitem{mina2014batman}
\BIBentryALTinterwordspacing
A.~X. Mina, ``{Batman, Pandaman and the Blind Man}: A case study in social change memes and internet censorship in {C}hina,'' \emph{Journal of Visual Culture}, vol.~13, no.~3, pp. 359--375, 2014. [Online]. Available: \url{https://doi.org/10.1177/1470412914546576}
\BIBentrySTDinterwordspacing

\bibitem{pramanick-etal-2021-detecting}
S.~Pramanick, D.~Dimitrov, R.~Mukherjee \emph{et~al.}, ``Detecting harmful memes and their targets,'' in \emph{Findings of ACL}, ser. ACL-IJCNLP~'21, aug 2021, pp. 2783--2796.

\bibitem{sharma2022findings}
S.~Sharma, T.~Suresh, A.~Kulkarni \emph{et~al.}, ``Findings of the constraint 2022 shared task on detecting the hero, the villain, and the victim in memes,'' in \emph{Proceedings of the Workshop on Combating Online Hostile Posts in Regional Languages during Emergency Situations}, 2022, pp. 1--11.

\bibitem{clef2024sharedtask}
A.~Barr{\'o}n-Cede{\~{n}}o, F.~Alam, T.~Chakraborty, T.~Elsayed, P.~Nakov, P.~Przyby{\l}a, J.~M. Stru{\ss}, F.~Haouari, M.~Hasanain, F.~Ruggeri, X.~Song, and R.~Suwaileh, ``The clef-2024 checkthat! lab: Check-worthiness, subjectivity, persuasion, roles, authorities, and adversarial robustness,'' in \emph{Advances in Information Retrieval}, N.~Goharian, N.~Tonellotto, Y.~He, A.~Lipani, G.~McDonald, C.~Macdonald, and I.~Ounis, Eds.\hskip 1em plus 0.5em minus 0.4em\relax Cham: Springer Nature Switzerland, 2024, pp. 449--458.

\bibitem{sharma-etal-2020-semeval}
\BIBentryALTinterwordspacing
C.~Sharma \emph{et~al.}, ``{S}em{E}val-2020 task 8: Memotion analysis- the visuo-lingual metaphor!'' in \emph{Proceedings of the Fourteenth Workshop on Semantic Evaluation}, ser. SemEval~'20, 2020, pp. 759--773. [Online]. Available: \url{https://www.aclweb.org/anthology/2020.semeval-1.99}
\BIBentrySTDinterwordspacing

\bibitem{kumar2019sarc}
A.~Kumar and G.~Garg, ``{Sarc-M}: Sarcasm detection in typo-graphic memes,'' in \emph{Proceedings of the International Conference on Advances in Engineering Science Management \& Technology}, ser. ICAESMT~'19, Dehradun, India, 2019.

\bibitem{zhou2021multimodal}
Y.~Zhou, Z.~Chen, and H.~Yang, ``Multimodal learning for hateful memes detection,'' in \emph{Proceedings of the International Conference on Multimedia Expo Workshops}, ser. ICMEW~'21, 2021, pp. 1--6.

\bibitem{memeEmotTAC}
S.~Sharma, R.~S, M.~S. Akhtar, and T.~Chakraborty, ``Emotion-aware multimodal fusion for meme emotion detection,'' \emph{IEEE Transactions on Affective Computing}, pp. 1800--1811, 2024.

\bibitem{DISARM:2022}
S.~Sharma, M.~S. Akhtar, P.~Nakov, and T.~Chakraborty, ``{DISARM}: Detecting the victims targeted by harmful memes,'' in \emph{Findings of NAACL}, Seattle, Washington, USA, 2022.

\bibitem{sharma-etal-2023-characterizing}
\BIBentryALTinterwordspacing
S.~Sharma, A.~Kulkarni, T.~Suresh \emph{et~al.}, ``Characterizing the entities in harmful memes: Who is the hero, the villain, the victim?'' in \emph{Proceedings of the 17th Conference of the European Chapter of the Association for Computational Linguistics}, A.~Vlachos and I.~Augenstein, Eds.\hskip 1em plus 0.5em minus 0.4em\relax Dubrovnik, Croatia: Association for Computational Linguistics, May 2023, pp. 2149--2163. [Online]. Available: \url{https://aclanthology.org/2023.eacl-main.157/}
\BIBentrySTDinterwordspacing

\bibitem{semeval2025task10}
J.~Piskorski \emph{et~al.}, ``{SemEval}-2025 task 10: Multilingual characterization and extraction of narratives from online news,'' in \emph{Proceedings of the 19th International Workshop on Semantic Evaluation}, ser. SemEval 2025, Vienna, Austria, July 2025.

\bibitem{Logically:2022}
L.~Kun, J.~Bankoti, and D.~Kiskovski, ``Logically at the {CONSTRAINT} 2022: Multimodal role labelling,'' in \emph{Proceedings of the Workshop on Combating Online Hostile Posts in Regional Languages during Emergency Situations}, ser. CONSTRAINT~'22, Dublin, Ireland, 2022.

\bibitem{DD-TIG:2022}
Z.~Zhou, H.~Zhao, J.~Dong, J.~Gao, and X.~Liu, ``{DD-TIG at Constraint@ACL2022}: Multimodal understanding and reasoning for role labeling of entities in hateful memes,'' in \emph{Proceedings of the Workshop on Combating Online Hostile Posts in Regional Languages during Emergency Situations}, ser. CONSTRAINT~'22, Dublin, Ireland, 2022.

\bibitem{Survey:2022:Harmful:Memes}
S.~Sharma, F.~Alam, M.~S. Akhtar \emph{et~al.}, ``Detecting and understanding harmful memes: A survey,'' in \emph{Proceedings of the 31st International Joint Conference on Artificial Intelligence}, ser. IJCAI-ECAI~'22, Vienna, Austria, 2022.

\bibitem{hee-etal-2024-recent}
M.~S. Hee, S.~Sharma, R.~Cao, P.~Nandi, P.~Nakov, T.~Chakraborty, and R.~K.-W. Lee, ``Recent advances in online hate speech moderation: Multimodality and the role of large models,'' in \emph{Findings of the Association for Computational Linguistics: EMNLP 2024}, Y.~Al-Onaizan, M.~Bansal, and Y.-N. Chen, Eds., Miami, Florida, USA, Nov. 2024, pp. 4407--4419.

\bibitem{Relia_Li_Cook_Chunara_2019}
K.~Relia, Z.~Li, S.~H. Cook, and R.~Chunara, ``Race, ethnicity and national origin-based discrimination in social media and hate crimes across 100 u.s. cities,'' \emph{ICWSM}, vol.~13, no.~01, pp. 417--427, Jul. 2019.

\bibitem{chengtroll2017}
\BIBentryALTinterwordspacing
J.~Cheng, M.~Bernstein, C.~Danescu-Niculescu-Mizil, and J.~Leskovec, ``Anyone can become a troll: Causes of trolling behavior in online discussions,'' in \emph{Proceedings of the 2017 ACM Conference on Computer Supported Cooperative Work and Social Computing}, ser. CSCW '17, New York, NY, USA, 2017, p. 1217–1230. [Online]. Available: \url{https://doi.org/10.1145/2998181.2998213}
\BIBentrySTDinterwordspacing

\bibitem{IJCAI2020:propaganda:survey}
G.~Da~San~Martino, S.~Cresci, A.~Barr\'{o}n-Cede\~no, S.~Yu, R.~Di~Pietro, and P.~Nakov, ``A survey on computational propaganda detection,'' in \emph{IJCAI}, Yokohama, Japan, 2020, pp. 4826--4832.

\bibitem{Gautam_Mathur_Gosangi_Mahata_Sawhney_Shah_2020}
\BIBentryALTinterwordspacing
A.~Gautam, P.~Mathur, R.~Gosangi, D.~Mahata, R.~Sawhney, and R.~R. Shah, ``{\#MeTooMA}: {M}ulti-aspect annotations of tweets related to the {MeToo} movement,'' \emph{Proceedings of the International AAAI Conference on Web and Social Media}, vol.~14, no.~1, pp. 209--216, May 2020. [Online]. Available: \url{https://ojs.aaai.org/index.php/ICWSM/article/view/7292}
\BIBentrySTDinterwordspacing

\bibitem{ousidhoum-etal-2019-multilingual}
\BIBentryALTinterwordspacing
N.~Ousidhoum, Z.~Lin, H.~Zhang, Y.~Song, and D.-Y. Yeung, ``Multilingual and multi-aspect hate speech analysis,'' in \emph{Proceedings of the 2019 Conference on Empirical Methods in Natural Language Processing and the 9th International Joint Conference on Natural Language Processing (EMNLP-IJCNLP)}, Hong Kong, China, Nov. 2019, pp. 4675--4684. [Online]. Available: \url{https://aclanthology.org/D19-1474}
\BIBentrySTDinterwordspacing

\bibitem{zain2018neural}
N.~Zainuddin, A.~Selamat, and R.~Ibrahim, ``Evaluating aspect-based sentiment classification on {T}witter hate speech using neural networks and word embedding features,'' in \emph{New Trends in Intelligent Software Methodologies, Tools and Techniques}, 2018, pp. 723--734.

\bibitem{shvets-etal-2021-targets}
\BIBentryALTinterwordspacing
A.~Shvets, P.~Fortuna, J.~Soler, and L.~Wanner, ``Targets and aspects in social media hate speech,'' in \emph{Proceedings of the 5th Workshop on Online Abuse and Harms (WOAH 2021)}, Online, Aug. 2021, pp. 179--190. [Online]. Available: \url{https://aclanthology.org/2021.woah-1.19}
\BIBentrySTDinterwordspacing

\bibitem{mathew2020hatexplain}
\BIBentryALTinterwordspacing
B.~Mathew \emph{et~al.}, ``Hatexplain: A benchmark dataset for explainable hate speech detection,'' \emph{Proceedings of the AAAI Conference on Artificial Intelligence}, vol.~35, no.~17, pp. 14\,867--14\,875, May 2021. [Online]. Available: \url{https://ojs.aaai.org/index.php/AAAI/article/view/17745}
\BIBentrySTDinterwordspacing

\bibitem{sap-etal-2020-social}
\BIBentryALTinterwordspacing
M.~Sap, S.~Gabriel, L.~Qin, D.~Jurafsky, N.~A. Smith, and Y.~Choi, ``Social bias frames: Reasoning about social and power implications of language,'' in \emph{Proceedings of the 58th Annual Meeting of the Association for Computational Linguistics}, ser. ACL~'20, Online, Jul. 2020, pp. 5477--5490. [Online]. Available: \url{https://aclanthology.org/2020.acl-main.486}
\BIBentrySTDinterwordspacing

\bibitem{ma-etal-2018-joint}
\BIBentryALTinterwordspacing
D.~Ma, S.~Li, and H.~Wang, ``Joint learning for targeted sentiment analysis,'' in \emph{Proceedings of the 2018 Conference on Empirical Methods in Natural Language Processing}, Brussels, Belgium, Oct.-Nov. 2018, pp. 4737--4742. [Online]. Available: \url{https://aclanthology.org/D18-1504}
\BIBentrySTDinterwordspacing

\bibitem{mitchell-etal-2013-open}
\BIBentryALTinterwordspacing
M.~Mitchell, J.~Aguilar, T.~Wilson, and B.~Van~Durme, ``Open domain targeted sentiment,'' in \emph{Proceedings of the 2013 Conference on Empirical Methods in Natural Language Processing}, Seattle, Washington, USA, Oct. 2013, pp. 1643--1654. [Online]. Available: \url{https://aclanthology.org/D13-1171}
\BIBentrySTDinterwordspacing

\bibitem{Zannettou2018}
\BIBentryALTinterwordspacing
S.~Zannettou, T.~Caulfield, J.~Blackburn \emph{et~al.}, ``On the origins of memes by means of fringe web communities,'' in \emph{Proceedings of the Internet Measurement Conference 2018}, ser. IMC '18, New York, NY, USA, 2018, p. 188–202. [Online]. Available: \url{https://doi.org/10.1145/3278532.3278550}
\BIBentrySTDinterwordspacing

\bibitem{CURwebist20}
C.~Sharma. and V.~Pulabaigari., ``A curious case of meme detection: An investigative study,'' in \emph{WEBIST}, 2020, pp. 327--338.

\bibitem{memenonmeme}
\BIBentryALTinterwordspacing
C.~Sharma, V.~Pulabaigari, and A.~Das, ``Meme vs. non-meme classification using visuo-linguistic association,'' in \emph{International Conference on Web Information Systems and Technologies}, 2020. [Online]. Available: \url{https://api.semanticscholar.org/CorpusID:227129761}
\BIBentrySTDinterwordspacing

\bibitem{suryawanshi-etal-2020-multimodal}
S.~Suryawanshi, B.~R. Chakravarthi, M.~Arcan, and P.~Buitelaar, ``\BIBforeignlanguage{English}{Multimodal meme dataset ({M}ulti{OFF}) for identifying offensive content in image and text},'' in \emph{\BIBforeignlanguage{English}{Proceedings of the Second Workshop on Trolling, Aggression and Cyberbullying}}, May 2020, pp. 32--41.

\bibitem{gomez2019exploring}
R.~Gomez, J.~Gibert, L.~Gomez, and D.~Karatzas, ``Exploring hate speech detection in multimodal publications,'' in \emph{Proceedings of the 2020 IEEE Winter Conference on Applications of Computer Vision}, ser. WACV~'20, 2020, pp. 1459--1467.

\bibitem{pramanick-etal-2021-momenta-multimodal}
S.~Pramanick, S.~Sharma, D.~Dimitrov \emph{et~al.}, ``{MOMENTA}: A multimodal framework for detecting harmful memes and their targets,'' in \emph{Findings of EMNLP 2021}, Nov. 2021, pp. 4439--4455.

\bibitem{9582340}
L.~Shang, C.~Youn, Y.~Zha, Y.~Zhang, and D.~Wang, ``{KnowMeme}: A knowledge-enriched graph neural network solution to offensive meme detection,'' in \emph{Proceedings of the 2021 IEEE 17th International Conference on eScience}, ser. eScience~'21, 2021, pp. 186--195.

\bibitem{karkkainen2019fairface}
K.~Karkkainen and J.~Joo, ``Fairface: Face attribute dataset for balanced race, gender, and age for bias measurement and mitigation,'' in \emph{WACV}, 2021, pp. 1548--1558.

\bibitem{sharma-etal-2023-memex}
S.~Sharma, R.~S, U.~Arora, M.~S. Akhtar, and T.~Chakraborty, ``{MEMEX}: Detecting explanatory evidence for memes via knowledge-enriched contextualization,'' in \emph{Proceedings of the 61st Annual Meeting of the Association for Computational Linguistics (Volume 1: Long Papers)}, Jul. 2023, pp. 5272--5290.

\bibitem{agarwal-etal-2024-mememqa}
S.~Agarwal, S.~Sharma, P.~Nakov, and T.~Chakraborty, ``{M}eme{MQA}: Multimodal question answering for memes via rationale-based inferencing,'' in \emph{Findings of the Association for Computational Linguistics: ACL 2024}, L.-W. Ku, A.~Martins, and V.~Srikumar, Eds.\hskip 1em plus 0.5em minus 0.4em\relax Bangkok, Thailand: Association for Computational Linguistics, Aug. 2024, pp. 5042--5078.

\bibitem{khanuja-etal-2020-gluecos}
S.~Khanuja, S.~Dandapat, A.~Srinivasan \emph{et~al.}, ``{GLUEC}o{S}: An evaluation benchmark for code-switched {NLP},'' in \emph{ACL '20}, Online, Jul. 2020, pp. 3575--3585.

\bibitem{khanuja2021muril}
S.~Khanuja, D.~Bansal, S.~Mehtani \emph{et~al.}, ``Muril: Multilingual representations for indian languages,'' 2021.

\bibitem{devlin-etal-2019-bert}
J.~Devlin \emph{et~al.}, ``{BERT}: Pre-training of deep bidirectional transformers for language understanding,'' in \emph{NAACL '19'}, 2019, pp. 4171--4186.

\bibitem{DBLP:journals/corr/abs-1810-04805}
\BIBentryALTinterwordspacing
J.~Devlin, M.~Chang, K.~Lee, and K.~Toutanova, ``{BERT:} pre-training of deep bidirectional transformers for language understanding,'' \emph{CoRR}, vol. abs/1810.04805, 2018. [Online]. Available: \url{http://arxiv.org/abs/1810.04805}
\BIBentrySTDinterwordspacing

\bibitem{das2022data}
M.~Das, S.~Banerjee, and A.~Mukherjee, ``Data bootstrapping approaches to improve low-resource abusive language detection for indic languages,'' \emph{arXiv preprint arXiv:2204.12543}, 2022.

\bibitem{grattafiori2024llama3herdmodels}
\BIBentryALTinterwordspacing
A.~Grattafiori \emph{et~al.}, ``The llama 3 herd of models,'' 2024. [Online]. Available: \url{https://arxiv.org/abs/2407.21783}
\BIBentrySTDinterwordspacing

\bibitem{jiang2023mistral7b}
\BIBentryALTinterwordspacing
A.~Q. Jiang, A.~Sablayrolles, A.~Mensch \emph{et~al.}, ``Mistral 7b,'' 2023. [Online]. Available: \url{https://arxiv.org/abs/2310.06825}
\BIBentrySTDinterwordspacing

\bibitem{qwen2.5}
\BIBentryALTinterwordspacing
Q.~Team, ``Qwen2.5: A party of foundation models,'' September 2024. [Online]. Available: \url{https://qwenlm.github.io/blog/qwen2.5/}
\BIBentrySTDinterwordspacing

\bibitem{he2021debertav3}
P.~He, J.~Gao, and W.~Chen, ``Debertav3: Improving deberta using electra-style pre-training with gradient-disentangled embedding sharing,'' 2021.

\bibitem{Qwen2VL}
P.~Wang, S.~Bai, S.~Tan \emph{et~al.}, ``Qwen2-vl: Enhancing vision-language model's perception of the world at any resolution,'' \emph{arXiv preprint arXiv:2409.12191}, 2024.

\bibitem{qwen2.5-VL}
\BIBentryALTinterwordspacing
Q.~Team, ``Qwen2.5-vl,'' January 2025. [Online]. Available: \url{https://qwenlm.github.io/blog/qwen2.5-vl/}
\BIBentrySTDinterwordspacing

\bibitem{liu2024llavanext}
\BIBentryALTinterwordspacing
H.~Liu \emph{et~al.}, ``Llava-next: Improved reasoning, ocr, and world knowledge,'' January 2024. [Online]. Available: \url{https://llava-vl.github.io/blog/2024-01-30-llava-next/}
\BIBentrySTDinterwordspacing

\bibitem{radford2021learningtransferablevisualmodels}
A.~Radford \emph{et~al.}, ``Learning transferable visual models from natural language supervision,'' in \emph{ICML}, 2021, pp. 8748--8763.

\end{thebibliography}



\end{document}